\documentclass{article}

\usepackage{microtype}
\usepackage{graphicx}
\usepackage{subfigure}
\usepackage{booktabs} 
\usepackage{caption}

\usepackage{times}
\usepackage{epsfig}
\usepackage{pifont}
\usepackage{diagbox}
\usepackage{multirow}
\usepackage{enumitem}
\usepackage{dblfloatfix}
\usepackage{scalerel}
\newcommand\mathbox[1]{\mathord{\ThisStyle{%
			\fboxsep3\LMpt\relax\kern1\LMpt\fbox{$\SavedStyle#1$}\kern1\LMpt}}}
		
\usepackage{hyperref}

\usepackage{color}
\usepackage[noend]{algorithmic}
\renewcommand{\algorithmicrequire}{\textbf{Input:}} 
\renewcommand{\algorithmicensure}{\textbf{Output:}} 

\usepackage{amsmath,amsthm,amssymb}
\usepackage{color}

\newcommand{\beq}{\begin{equation}}
\newcommand{\eeq}{\end{equation}}
\newcommand{\beqs}{\begin{eqnarray}}
\newcommand{\eeqs}{\end{eqnarray}}
\newcommand{\barr}{\begin{array}}
	\newcommand{\earr}{\end{array}}
\newcommand{\bali}{\begin{aligned}}
	\newcommand{\eali}{\end{aligned}}

\newcommand{\Ebb}[0]{\ensuremath{\mathbb{E}} }

\newcommand{\Rbb}[0]{\ensuremath{\mathbb{R}} }

\newcommand{\ie}[0]{\emph{i.e., }}

\newcommand{\eg}[0]{\emph{e.g., }}

\newcommand{\Wten}[0]{\ensuremath{{\boldsymbol{\mathsf{W}}}} }

\newcommand{\Mmat}[0]{\ensuremath{{\bf M}} }

\newcommand{\bds}[1]{\boldsymbol{#1}}

\newcommand{\sv}[0]{\ensuremath{\boldsymbol{s}} }
\newcommand{\tv}[0]{\ensuremath{\boldsymbol{t}} }

\newcommand{\xv}[0]{\ensuremath{\boldsymbol{x}} }
\newcommand{\yv}[0]{\ensuremath{\boldsymbol{y}} }
\newcommand{\zv}[0]{\ensuremath{\boldsymbol{z}} }

\newcommand{\betav}[0]{\ensuremath{\boldsymbol{\beta}} }
\newcommand{\gammav}[0]{\ensuremath{\boldsymbol{\gamma}} }

\newcommand{\etav}[0]{\ensuremath{\boldsymbol{\eta}} }

\usepackage[accepted]{icml2020}

\icmltitlerunning{On Leveraging Pretrained GANs for Generation with Limited Data}

\begin{document}

\twocolumn[
\icmltitle{On Leveraging Pretrained GANs for Generation with Limited Data\\
}
\icmlsetsymbol{equal}{*}

\begin{icmlauthorlist}
\icmlauthor{Miaoyun Zhao}{equal,to}
\icmlauthor{Yulai Cong}{equal,to}
\icmlauthor{Lawrence Carin}{to}
\end{icmlauthorlist}

\icmlaffiliation{to}{Department of Electrical and Computer Engineering, Duke University, Durham NC, USA}

\icmlcorrespondingauthor{Miaoyun Zhao}{miaoyun9zhao@gmail.com}
\icmlcorrespondingauthor{Yulai Cong}{yulaicong@gmail.com}

\icmlkeywords{GAN, Transfer learing, limited data}

\vskip 0.3in
]

\printAffiliationsAndNotice{\icmlEqualContribution} 

\begin{abstract}
	Recent work has shown generative adversarial networks (GANs) can generate highly realistic images, that are often indistinguishable (by humans) from real images. Most images so generated are not contained in the training dataset, suggesting potential for augmenting training sets with GAN-generated data. 
	While this scenario is of particular relevance when there are limited data available, there is still the issue of training the GAN itself based on that limited data. 
	To facilitate this, we leverage existing GAN models pretrained on large-scale datasets (like ImageNet) to introduce additional knowledge (which may not exist within the limited data), following the concept of transfer learning.
	Demonstrated by natural-image generation, we reveal that low-level filters (those close to observations) of both the generator and discriminator of pretrained GANs can be transferred to facilitate generation in a perceptually-distinct target domain with limited training data.  
	To further adapt the transferred filters to the target domain, we propose adaptive filter modulation (AdaFM). 
	An extensive set of experiments is presented to demonstrate the effectiveness of the proposed techniques on generation with limited data.

	
\end{abstract}


\section{Introduction}

Recent research has demonstrated the increasing power of generative adversarial networks (GANs) to generate high-quality samples, that are often  indistinguishable from real data \cite{karras2017progressive,GoogleCompareGAN,miyato2018spectral,brock2018large,Karras_2019_CVPR}; this demonstrates the capability of GANs to exploit the valuable information within the underlying data distribution.
Although many powerful GAN models pretrained on large-scale datasets have been released, few efforts have been made \cite{giacomello2019transfer} to take advantage of the valuable information within those models to facilitate downstream tasks; this shows a clear contrast with the popularity of transfer learning for discriminative tasks (\eg to reuse the feature extractor of a pretrained classifier) \cite{bengio2012deep,donahue2014decaf,luo2017label,zamir2018taskonomy},
and transfer learning in natural language processing (\eg to reuse the expensively-pretrained BERT model) \cite{devlin2018bert,bao2019transfer,peng2019transfer,mozafari2019bert}. 

Motivated by the significant value of released pretrained GAN models, we propose to leverage the information therein to facilitate downstream tasks in a target domain with limited training data. This situation arises frequently due to expensive data collection or privacy issues that may arise in medical or biological applications \cite{yi2019generative}. 
We concentrate on the challenging scenario of GAN model development when limited training data are available.
One key observation motivating our method is that a well-trained GAN can generate realistic images not observed in the training dataset \cite{brock2018large,Karras_2019_CVPR,han2019learning}, demonstrating the generalization ability of GANs to capture the training data manifold.
Likely arising from novel combinations of information/attributes/styles (see stunning illustrations in StyleGAN \cite{Karras_2019_CVPR}), this generalization of GANs is extremely appealing for scenarios in which there are limited data \cite{yi2019generative,han2019learning}. For example, GANs can be used to augment the training set via realistic data generation, to alleviate overfitting or provide regularizations for classification \cite{wang2017effectiveness,frid2018gan}, segmentation \cite{bowles2018gan}, or detection \cite{han2019learning, han2020infinite}.

However, the limited data in the target domain manifests a problem in learning the underlying GAN model, as GANs typically require substantial training data. 
When a limited quantity of data are available, to naively train a GAN is prone to overfitting, as powerful GAN models have numerous parameters that are essential for realistic generation \cite{bermudez2018learning,bowles2018gan,frid2018gan,finlayson2018towards}. 
To alleviate overfitting, one may consider transferring additional information from other domains via transfer learning, which may deliver simultaneously better training efficiency and performance \cite{caruana1995learning,bengio2012deep,sermanet2013overfeat, donahue2014decaf,zeiler2014visualizing,girshick2014rich}.
However, most transfer learning work has focused on discriminative tasks, based on the foundation that low-level filters (those close to \emph{input} observations) of a classifier pretrained on a large-scale source dataset are fairly general (like Gabor filters) and thus transferable to different target domains \cite{yosinski2014transferable}; as the well-trained low-level filters (often data-demanding \cite{fregier2019mind2mind,noguchi2019image}) provide additional information, transfer learning often leads to better performance \cite{yosinski2014transferable,long2015learning,noguchi2019image}.
Compared to transfer learning on discriminative tasks, fewer efforts have been made for generation tasks \cite{shin2016generative,wang2018transferring,noguchi2019image}, as summarized in Section \ref{sec:related}.
The work presented here addresses this challenge, considering transfer learning for GANs when there are limited data in the target domain.

Leveraging insights from the aforementioned transfer learning on discriminative tasks, we posit that the low-level filters of a GAN discriminator pretrained on a large-scale source dataset are likely to be generalizable and hence transferable to various target domains. 
For a pretrained GAN generator, it's shown \cite{bau2017network,bau2018gan,Karras_2019_CVPR} that the low-level layers (those close to \emph{output} observations) capture properties of generally-applicable local patterns like materials, edges, and colors, while the high-level layers (those distant from observations) are associated with more domain-specific semantic aspects of data. 
We therefore consider transferring/freezing the low-level filters from both the generator and discriminator of a pretrained GAN model to facilitate generation in perceptually-distinct target domains with limited training data.
As an illustrative example, we consider the widely studied GAN scenario of natural-image generation, although the proposed techniques are general and may be applicable to other domains, such as in medicine or biology. 
The principal contributions of this paper are as follows.  
\begin{itemize}[leftmargin=*]
	\vspace{-\topsep}
	\setlength{\itemsep}{1pt}
	\setlength{\parskip}{0.1pt}
	\setlength{\parsep}{0.1pt}
	
	\item We demonstrate empirically that the low-level filters (within both the generator and the discriminator) of a GAN model, pretrained on large-scale datasets, can be transferred to perceptually-distinct target domains, yielding improved GAN performance in scenarios for which limited training data are available.
	
	\item We tailor a compact domain-specific network to harmoniously cooperate with the transferred low-level filters, which enables style mixing for diverse synthesis. 
	
	\item To better adapt the transferred filters to the target domain, we introduce an approach we term adaptive filter modulation (AdaFM), that delivers boosted performance.
	
	\item Extensive experiments are conducted to verify the effectiveness of the proposed techniques.
	\vspace{-0.3 cm}
\end{itemize}

\section{Background and Related Work\label{sec:related}}

\subsection{Generative Adversarial Networks (GANs)}

GANs have demonstrated increasing power for synthesizing highly realistic data  \cite{brock2018large,Karras_2019_CVPR,karras2019analyzing}; 
accordingly, they are widely applied in various research fields, such as image \cite{hoffman2017cycada,ledig2017photo,ak2019attribute}, text \cite{lin2017adversarial,fedus2018maskgan,wang2018sentigan,wang2019open}, video \cite{mathieu2015deep,wang2017fast, wang2018video, chan2019everybody}, and audio \cite{engel2019gansynth,yamamoto2019probability,kumar2019melgan}.

A GAN often consists of two adversarial components, \ie a generator $G$ and a discriminator $D$. 
As the adversarial game proceeds, the generator learns to synthesize increasingly realistic fake data, to confuse the discriminator; the discriminator seeks to discriminate real and fake data synthesized by the generator.
The standard GAN objective \cite{goodfellow2014generative} is 
\begin{equation}\label{eq:standard_GAN_loss}
\bali
	\min_{G} \max_{D} \,\, & \Ebb_{\xv\sim q_{\text{data}}(\xv)} \big[\log D(\xv)\big] \\
	& + \Ebb_{\zv \sim p(\zv)} \big[\log (1-D(G(\zv))) \big],
\eali
\end{equation}
where $p(\zv)$ is an easy-to-sample distribution, like a normal distribution, and $q_{\text{data}}(\xv)$ is the empirical data distribution.

\vspace{-1mm}
\subsection{GANs on Limited Data}

Existing work addressing the design of GANs based on limited data can be roughly summarized into two groups.

\textbf{Exploit GANs for better usage of the information within the limited data.} 
In addition to traditional data augmentations like shift, zooming, rotation, or flipping, GANs trained on limited data can be leveraged for synthetic augmentations, like synthesized observations with transformed styles \cite{wang2017effectiveness} or fake observation-label/segmentation pairs \cite{bowles2018gan,frid2018gan,han2019learning,han2020infinite}. 
However, because of the limited available data, a relatively small GAN model is often employed, leading to reduced generative power. Furthermore, only the information within the limited data are utilized.

\begin{figure*}[htb]
	\begin{center}
		\includegraphics[width=1.86\columnwidth]{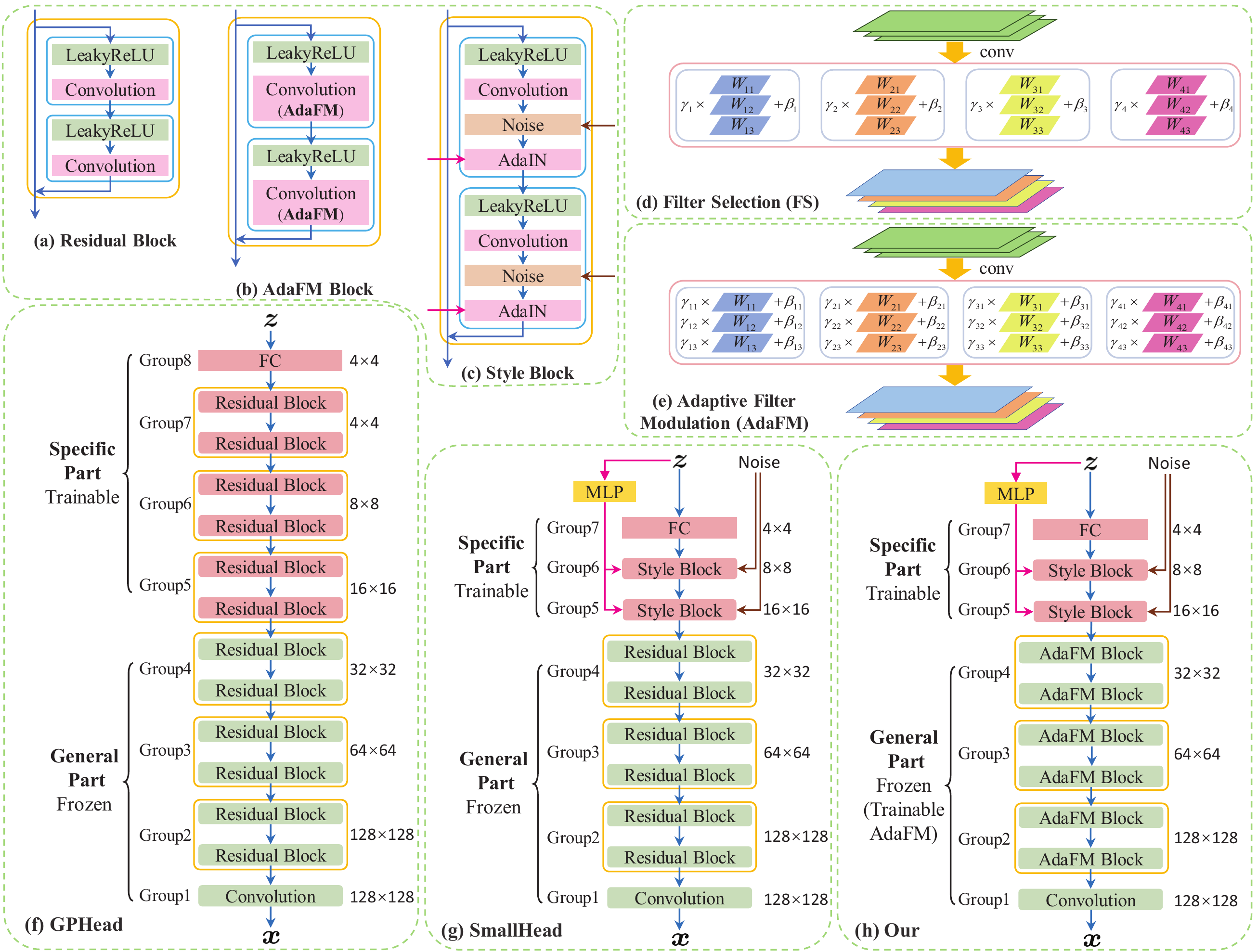}
		\vspace{-0.2cm}
		\caption{ 
			Network architectures.
			A ``group'' in general contains blocks with the same feature-map size. 
			The multilayered perceptron (MLP) consists of 8 fully-connected (FC) layers. With a 64-dimensional $\zv$, the number of trainable parameters is 79.7M for GPHead, 23.0M for SmallHead, and 24.4M for Our model.
		}
		\label{fig:Network_all}
	\end{center}
	\vspace{-0.5cm}
\end{figure*}

\textbf{Use GANs to transfer additional information to facilitate generation with limited data.} 
As the available data are limited, it's often preferred to transfer additional information from other domains via transfer learning \cite{yosinski2014transferable,long2015learning,noguchi2019image}.
TransferGAN \cite{wang2018transferring} initializes the target GAN model with parameters pretrained on source large-scale datasets, followed by fine-tuning the whole model with the limited target data. As source model architecture (often large) is directly transferred to the target domain, fine-tuning with too limited target data may suffer from overfitting, as verified empirically in our experiments; since the high-level semantically specific filters are also transferred, the similarity between the source and target domains is often critical for a beneficial transfer \cite{wang2018transferring}. 
Similarly, based on the assumption that the source and target domains share the same support, \citet{wang2020minegan} introduces an additional miner network to mine knowledge from pretrained GANs to form target generation, likewise fine-tuning the whole model with the limited data.

Different from the above fine-tuning methods, \citet{noguchi2019image} propose batch-statistics adaptation (BSA) to transfer/freeze the whole source generator but introduce new trainable parameters to adapt its hidden batch statistics for generation with extremely-limited target data; however, the generator is not adversarially trained (L1/Perceptual loss is used instead), leading to blurry generation in the target domain \cite{noguchi2019image}. 
By comparison, our method transfers/freezes the generally-applicable low-level filters -- in the generator and discriminator -- from source to (perceptually-distinct) target domains, followed by employing a small tailored high-level network and the newly-introduced adaptive filter modulation (AdaFM) to better adapt to the target limited data. 
Accordingly, our proposed method, when compared to the fine-tuning methods \cite{wang2018transferring, wang2020minegan}, is expected to suffer less from overfitting and behave more robustly in the presence of differences between the source and target domains; when compared to \cite{noguchi2019image}, our method is more flexible and provides clearly better generation, thanks to its adversarial training.

Recently, a concurrent work \cite{mo2020freeze} reveals that freezing low-level layers of the GAN discriminator delivers better fine-tuning of GANs; it may be viewed as a special case of our method, which transfers/freezes low-level layers of {\em both} the generator and discriminator, leveraging a tailored high-level network and employing the proposed AdaFM. Consequently, our method is expected to perform better on (extremely) limited data.

\vspace{-0.1 cm}
\section{Proposed Method}
\label{sec:proposed_method}

For GAN training with limited data in the target domain, we propose to transfer additional information by leveraging the valuable low-level filters (those close to observations) from existing GANs pretrained on large-scale source datasets. 
Combining that prior knowledge, from the transferred low-level filters that are often generally-applicable but data-demanding to train \cite{yosinski2014transferable,fregier2019mind2mind}, one may expect less overfitting and hence better GAN performance. 
Specifically, given a pretrained GAN model, we reuse its low-level filters (termed the {\em generalizable or general part} of the model) in a target domain, and replace the high-level layers (termed the domain {\em specific part}) with another smaller network, and then train that specific part using the limited target data, while keeping the transferred general part frozen (see Figures \ref{fig:Network_all}(f) and (g)). Hence, via this approach, we leverage the transferred general part, trained on a much larger source dataset, and by employing the simplified domain-specific part, the total number of parameters that need be learned is reduced substantially, aligning with the limited target-domain data.

In what follows, we take natural image generation as an example, and present our method by answering three questions in Sections \ref{sec:reuse_Big}, \ref{sec:build_small_net}, and \ref{sec:Ada_FM}, respectively:
\vspace{-0.15 cm}
\begin{itemize}[leftmargin=*,topsep=0.cm,itemsep=0.05cm,partopsep=0cm,parsep=0cm]
	\item How to specify the general part appropriate for transfer?
	\item How to tailor the specific part so that it is simplified?
	\item How to better adapt the transferred general part?
\end{itemize}

Before introducing the proposed techniques in detail, we first discuss source datasets, available pretrained GAN models, and evaluation metrics. 
Intuitively, to realize generally-applicable low-level filters, one desires a large-scale source dataset with rich diversity. In the context of image analysis, a common choice is the ImageNet dataset \cite{krizhevsky2012imagenet,shin2016deep}, which contains 1.2 million high-resolution images from 1000 classes; we consider this as the source dataset.
Concerning publicly available GAN models pretrained on ImageNet, available choices include SNGAN \cite{miyato2018spectral}, GP-GAN \cite{mescheder2018training}, and BigGAN \cite{brock2018large}; we select the pretrained GP-GAN model (with resolution $128 \times 128$) because of its well-written codebase.
To evaluate the generative performance, we adopt the widely used Fr{\'e}chet inception distance (FID, lower is better) \cite{heusel2017gans}, a metric assessing the realism and variation of generated samples \cite{zhang2018self}.

\vspace{-0.2cm}
\subsection{On Specifying the General Part for Transfer}
\label{sec:reuse_Big}
\vspace{-0.1cm}

\begin{figure}[!tb]
	\centering
		\includegraphics[width=\columnwidth]{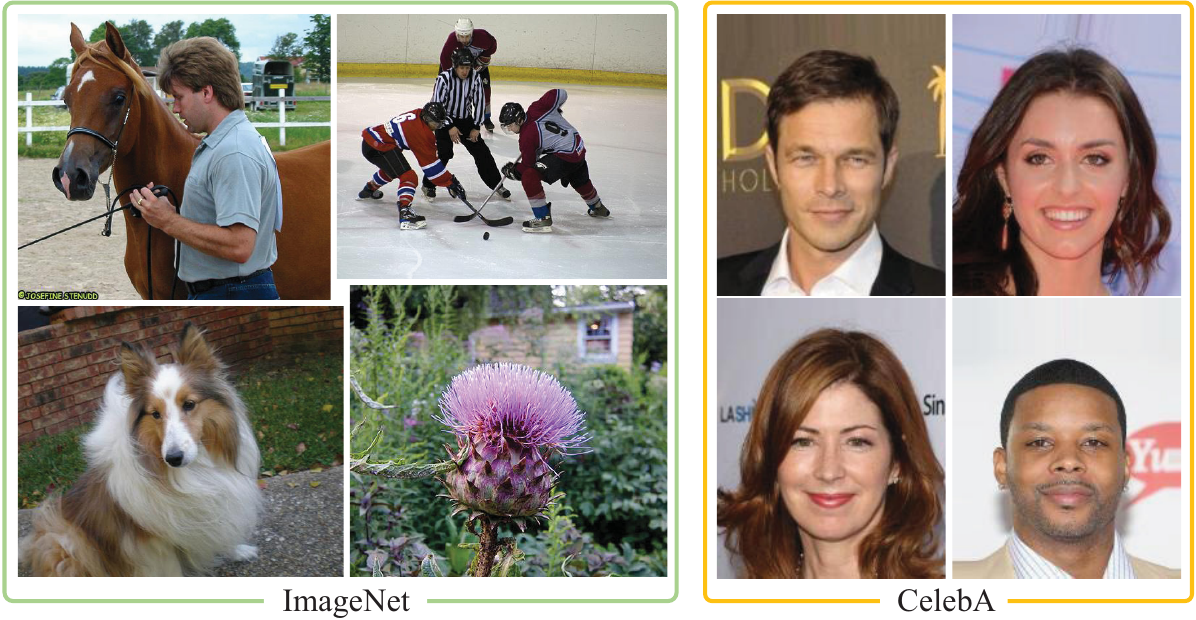}
		\vspace{-0.6cm}
		\caption{ 
			Sample images from the ImageNet and CelebA datasets. 
			Although quite different, they are likely to share the same set of low-level filters describing basic shapes, like lines, curves and textures.
		}
		\label{fig:original_ImageNet_CelebA}
	\vspace{-4 mm}
\end{figure}

As mentioned in the Introduction, both generative and discriminative image models share a similar pattern: higher-level convolutional filters portray more domain-specific semantic information, while lower-level filters portray more generally applicable information \cite{yosinski2014transferable,zeiler2014visualizing,bau2017network,bau2018gan}. 
Given the GP-GAN model pretrained on ImageNet, the question is how to specify the low-level filters (\ie the general part of the model) to be transferred to a target domain.
Generally speaking, the optimal solution is likely to be a compromise depending on the available target-domain data; if plenty of data are provided, less low-level filters should be transferred (less prior knowledge need be transferred), but when the target-domain data are limited, it's better to transfer more filters (leveraging more prior information).
We empirically address that question by transferring the pretrained GP-GAN model to the CelebA dataset \cite{liu2015deep}, which is fairly different from the source ImageNet (see Figure \ref{fig:original_ImageNet_CelebA}). 
It's worth emphasizing that the general part discovered here\footnote{
	This general part may not be the optimum with the best generalization, which is deemed intractable. 
	The key is that it's applicable to various target domains (see the experiments); in addition, the AdaFM technique introduced in Section \ref{sec:Ada_FM} delivers significantly improved adaptation/transfer and thus greatly relaxes the requirement of an optimal selection of the general part.
} also delivers excellent results on three other datasets (see the experiments and Appendix \ref{secapp:contri_AdaFM}).

\vspace{-0.1cm}
\subsubsection{General Part of the Generator}

To determine the appropriate general part of the GP-GAN generator, to be transferred to the target CelebA dataset,\footnote{
	To verify the generalization of the pretrained filters, we bypass the limited-data assumption in this section and employ the whole CelebA data for training.
} we employ the GP-GAN architecture and design experiments with increasing number of lower layers included in the transferred/frozen general part of the generator; the remaining specific part (not transferred) of the generator (see Figure \ref{fig:Network_all}(f)) and the discriminator are reinitialized and trained with CelebA.

\begin{figure}[tb]
	\centering
		\includegraphics[width=\columnwidth]{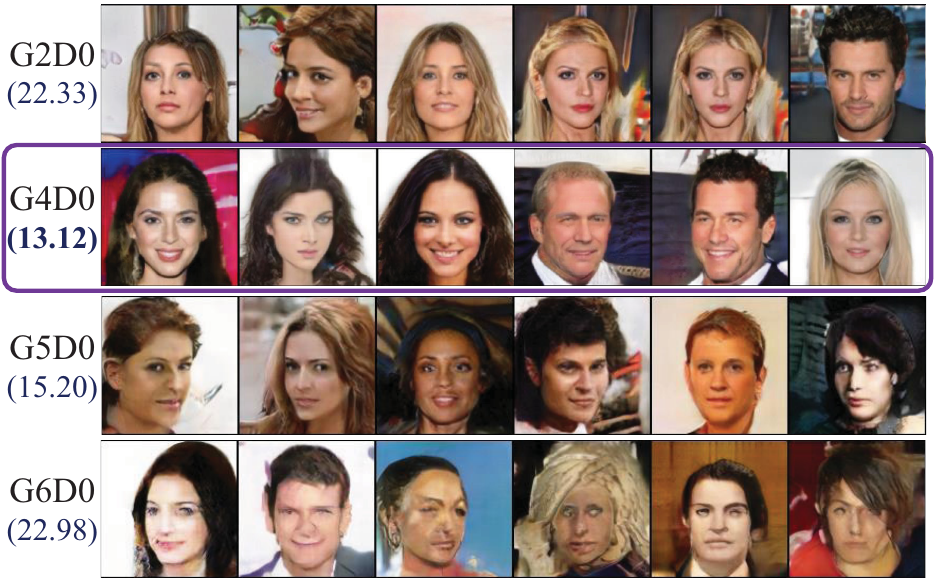}
		\vspace{-6 mm}
		\caption{Generated samples and FIDs from different settings for the general part of the generator.
			GmDn indicates freezing the lower $m/n$ groups as the general part of generator/discriminator.
		}
		\label{fig:fix_G_results}
	\vspace{-4 mm}
\end{figure}

Four settings for the generator general part are tested, \ie 2, 4, 5, and 6 lower groups to be transferred (termed G2, G4, G5, and G6, respectively; G4 is illustrated in Figure \ref{fig:Network_all}(f)).
After 60,000 training iterations (generative quality stabilizes by then), and we show in Figure \ref{fig:fix_G_results} the generated samples and FIDs of the four settings.
It's clear that transferring the G2/G4 generator general part delivers decent generative quality (see eye details, hair texture, and cheek smoothness), despite the fact that the source ImageNet is perceptually distant from the target CelebA, confirming the generalizable nature of the low-level filters within up to 4 lower groups of the pretrained GP-GAN generator (which is also verified on three other datasets in the experiments). The lower FID of G4 than that of G2 indicates that transferring more low-level filters pretrained on large-scale source datasets potentially benefits better performance in target domains.\footnote{
	Another reason might be that to train well-behaved low-level filters is time-consuming and data-demanding \cite{fregier2019mind2mind,noguchi2019image}.
	The worse FID of G2 is believed caused by the insufficiently trained low-level filters, as we find the images from G2 show a relatively lower diversity and contain strange textures in the details (see Figure \ref{fig:water} in Appendix). FID is biased toward texture rather than shape \cite{karras2019analyzing}.
}
But when we transfer and hence freeze more groups as the general part of the generator (\ie G5 and G6), the generative quality drops quickly; this is expected as higher-level filters are more specific to the source ImageNet and may not fit the target CelebA. 
By reviewing Figure \ref{fig:fix_G_results}, we choose G4 as the setting for the generator general part for transfer.

\vspace{-0.1cm}
\subsubsection{General Part of the Discriminator}
\vspace{-0.1cm}

Based on the G4 general part of the generator, we next conduct experiments to specify the general part of the discriminator. 
We consider transferring/freezing 0, 2, 3, and 4 lower groups of the pretrained GP-GAN discriminator (termed D0, D2, D3, and D4, respectively; D2 is illustrated in Figure \ref{fig:Framework_D} of the Appendix).
Figure \ref{fig:fix_D_results} shows the generated samples and FIDs for each setting.
Similar to what's observed for the generator, transferring low-level filters from the pretrained GP-GAN discriminator also benefits a better generative performance (compare the FID of D0 with that of D2), thanks to the additional information therein; however, as the higher-level filters are more specific to the source ImageNet, transferring them may lead to a decreased generative quality (see the results from D3 and D4).

Considering both the generator and discriminator, we transfer/freeze the G4D2 general part\footnote{
	Appendix \ref{sec:select_GmDn} discusses other settings for the general part.
} from the pretrained GP-GAN model, which will be shown in the experiments to work quite well on three other target datasets.

\begin{figure}[tb]
	\centering
		\includegraphics[width=\columnwidth]{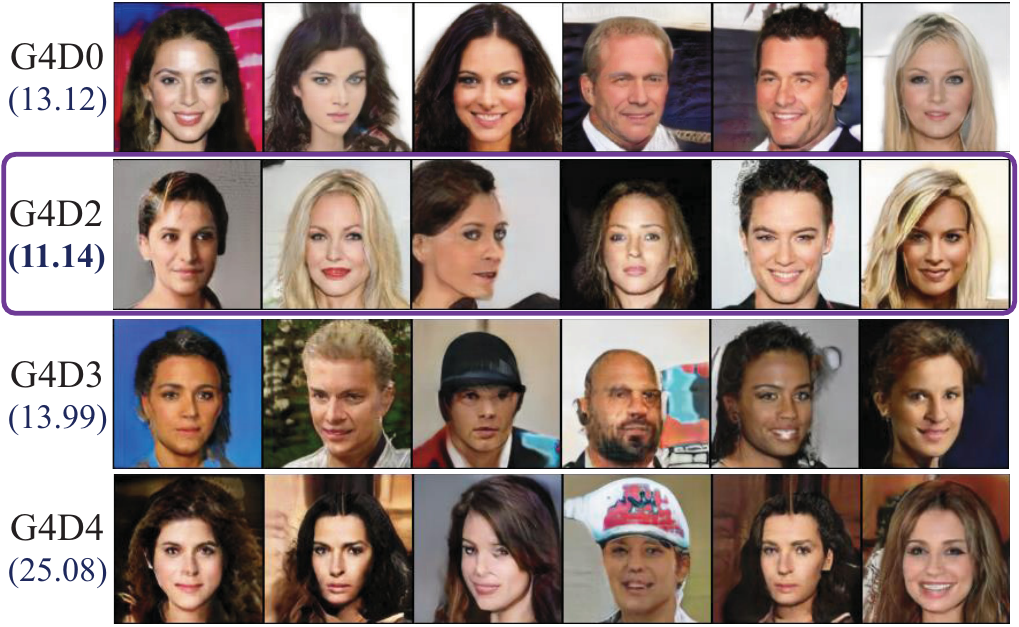}
		\vspace{-6 mm}
		\caption{ 
			Generated samples and FIDs from different settings for the general part of the discriminator.
		}
		\label{fig:fix_D_results}
	\vspace{-2 mm}
\end{figure}

\vspace{-0.1cm}
\subsection{On Tailoring the High-Level Specific Part}
\label{sec:build_small_net}

\begin{figure}[tb]
	\centering
		\includegraphics[width=\columnwidth]{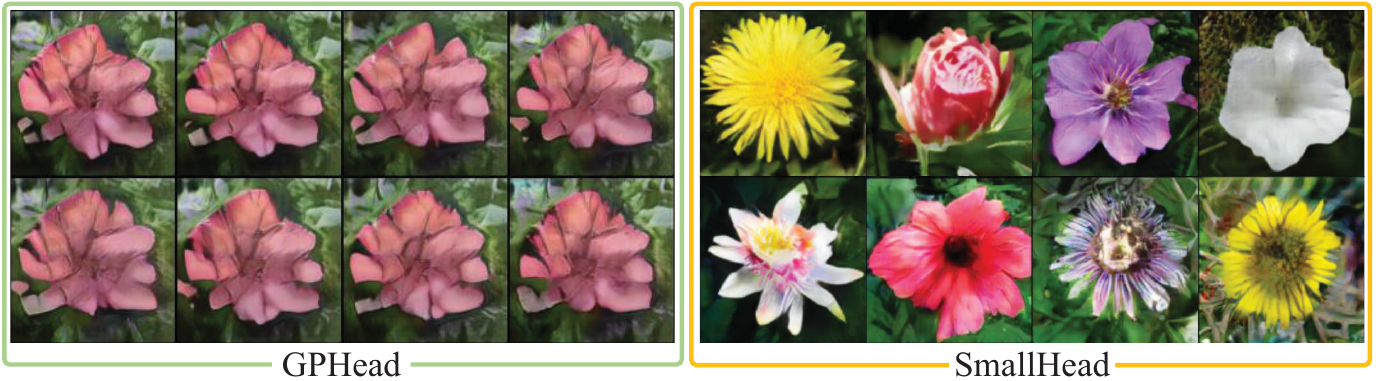}
		\vspace{-6 mm}
		\caption{ 
			Generated images from the GPHead and SmallHead trained on the Flowers dataset \cite{nilsback2008automated}.
		}
		\label{fig:G4D2_failed}
	\vspace{-4 mm}
\end{figure}

Even with the transferred/frozen G4D2 general part, the remaining specific (target-dependent) part may contain too many trainable parameters for the limited target-domain data (\eg the GPHead model in Figure \ref{fig:Network_all}(f) shows mode collapse (see Figure \ref{fig:G4D2_failed}) when trained on the small Flowers dataset \cite{nilsback2008automated}); another consideration is that, when using GANs for synthetic augmentation for applications with limited data, style mixing is a highly appealing capability \cite{wang2017effectiveness}.
Motivated by those considerations, we propose to replace the high-level specific part of GP-GAN with a tailored smaller network (see Figure \ref{fig:Network_all}(g)), to alleviate overfitting, enable style mixing, and also lower the computational/memory cost.

Specifically, that tailored specific part is constructed as a fully connected (FC) layer followed by two successive style blocks (borrowed from StyleGAN \cite{Karras_2019_CVPR} with an additional short cut, see Figure \ref{fig:Network_all}(c)). 
Similar to StyleGAN, the style blocks enable unsupervised disentanglement of high-level attributes, which may benefit an efficient exploration of the underlying data manifold and thus lead to better generation; they also enable generating samples with new attribute combinations (style mixing), which dramatically enlarges the generation diversity (see Figures \ref{fig:mix_style_Flowers} and \ref{fig:mix_style_CELEBA} of the Appendix). 
We term the model consisting of the tailored specific part and the G4D2 general part as SmallHead. 
Note that proposed specific part is also used in our method (see Figure \ref{fig:Network_all}(h)).
Different from the GPHead, the SmallHead has proven to train in a stable manner, without mode collapse on Flowers (see Figure \ref{fig:G4D2_failed}).
In the experiments, the SmallHead is found to work well on other small datasets.

\vspace{-1mm}
\subsection{Better Adaption of the Transferred General Part}
\label{sec:Ada_FM}

\begin{figure}[tb]
	\centering
	\includegraphics[width=\columnwidth]{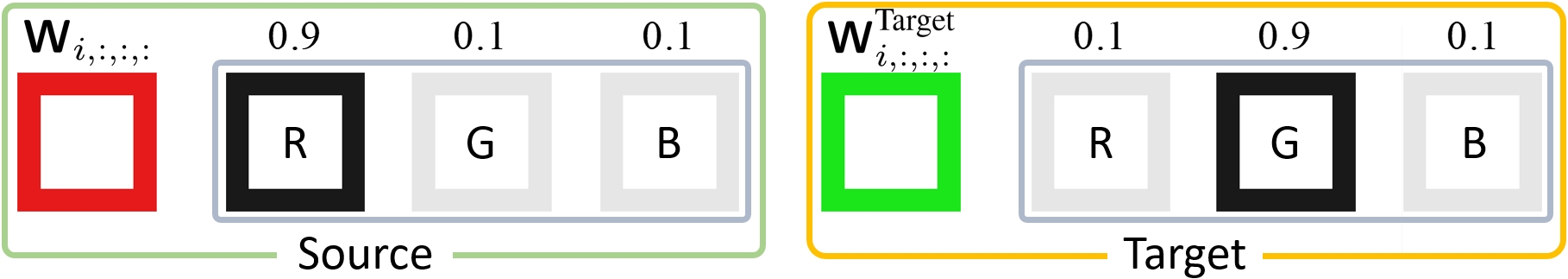}
	\vspace{-6mm}
	\caption{ 
		An illustrative example motivating AdaFM. Both the source and target domains share the same basic shape within each channel but use a different among-channel correlation.
		AdaFM learns $\gammav_{i,:}=[1/9,9,1]$ to adapt source $\Wten_{i,:,:,:}$ to target $\Wten_{i,:,:,:}^{\text{Target}}$.
	}
	\label{fig:AdaFM_apple}
	\vspace{-3 mm}
\end{figure}

Based on the above transferred general part and tailored specific part, we next present a new technique, termed adaptive filter modulation (AdaFM), to better adapt the transferred low-level filters to target domains for boosted performance, as shown in the experiments. In this proposed approach, we no longer just ``freeze'' the transferred filters upon transfer, but rather augment them in a target-dependent manner.

Motivated by the style-transfer literature \cite{huang2017arbitrary,noguchi2019image}, where one manipulates the style of an image by modifying the statistics ($e.g.$, mean or variance) of its latent feature maps, we alternatively consider the variant of manipulating the style of a function (\ie the transferred general part) by modifying the statistics of its convolutional filters via AdaFM.

Specifically, given a transferred convolutional filter $\Wten \in\Rbb^{C_{\text{out}} \times C_{\text{in}} \times K_1 \times K_2}$ in the general part, where $C_{\text{in}}/C_{\text{out}}$ denotes the number of input/output channels and $K_1 \times K_2$ is the kernel size, AdaFM introduces a small amount of learnable parameters, \ie scale $\gammav \in \Rbb^{C_{\text{out}} \times C_{\text{in}}}$ and shift $\betav \in \Rbb^{C_{\text{out}} \times C_{\text{in}}}$, to modulate its statistics via 
\begin{equation}\label{eq:AdaFM}
\begin{aligned}
\Wten^{\text{AdaFM}}_{i,j,:,:} = \gammav_{i,j} \Wten_{i,j,:,:} + \betav_{i,j},
\end{aligned}
\end{equation}
for $i \in \{1,2,\cdots, C_{\text{out}}\}$ and $j \in \{1,2,\cdots, C_{\text{in}}\}$.
$\Wten^{\text{AdaFM}}$ is then used to convolve with input feature maps for output ones.
Applying AdaFM to convolutional kernels of a residual block (see Figure \ref{fig:Network_all}(a)) \cite{he2016deep} gives the AdaFM block (see Figure \ref{fig:Network_all}(b)). 
With the residual blocks of the SmallHead replaced with AdaFM blocks, we yield our generator, as shown in Figure \ref{fig:Network_all}(h), which delivers boosted performance than the SmallHead in the experiments.

For better understanding the power of our AdaFM, below we draw parallel connections to two related techniques, which may be viewed as the special cases of AdaFM to some extent. 
The first one is the weight demodulation revealed in the recent StyleGAN2 \cite{karras2019analyzing}, a model with state-of-the-art generative performance.
Compared with AdaFM, the weight demodulation employs zero shift $\betav = 0$ and a rank-one scale $\gammav=\etav \sv^T$, where style $\sv \in \Rbb^{C_{\text{in}}}$ is produced by a trainable mapping network (often a MLP) and $\etav \in \Rbb^{C_{\text{out}}}$ is calculated as 
\beq\label{eq:eta_WD}
\setlength\abovedisplayskip{3.0pt}
\setlength\belowdisplayskip{3.0pt}
	\etav_i = \frac{1}{\sqrt{\epsilon + \sum_{j, k_1, k_2} \sv_j \Wten_{i,j,k_{1},k_{2}}}},
\eeq
where $\epsilon$ is a small constant to avoid numerical issues.
Despite being closely related, AdaFM and the weight demodulation are motivated differently. We propose AdaFM to better adapt the transferred filters to target domains, while the weight demodulation is used to relax instance normalization while keeping the capacity for controllable style mixing  \cite{karras2019analyzing}. 
See Appendix \ref{appsec:compare_WD} for more discussions.

\begin{figure}
	\centering
	\includegraphics[width=\columnwidth]{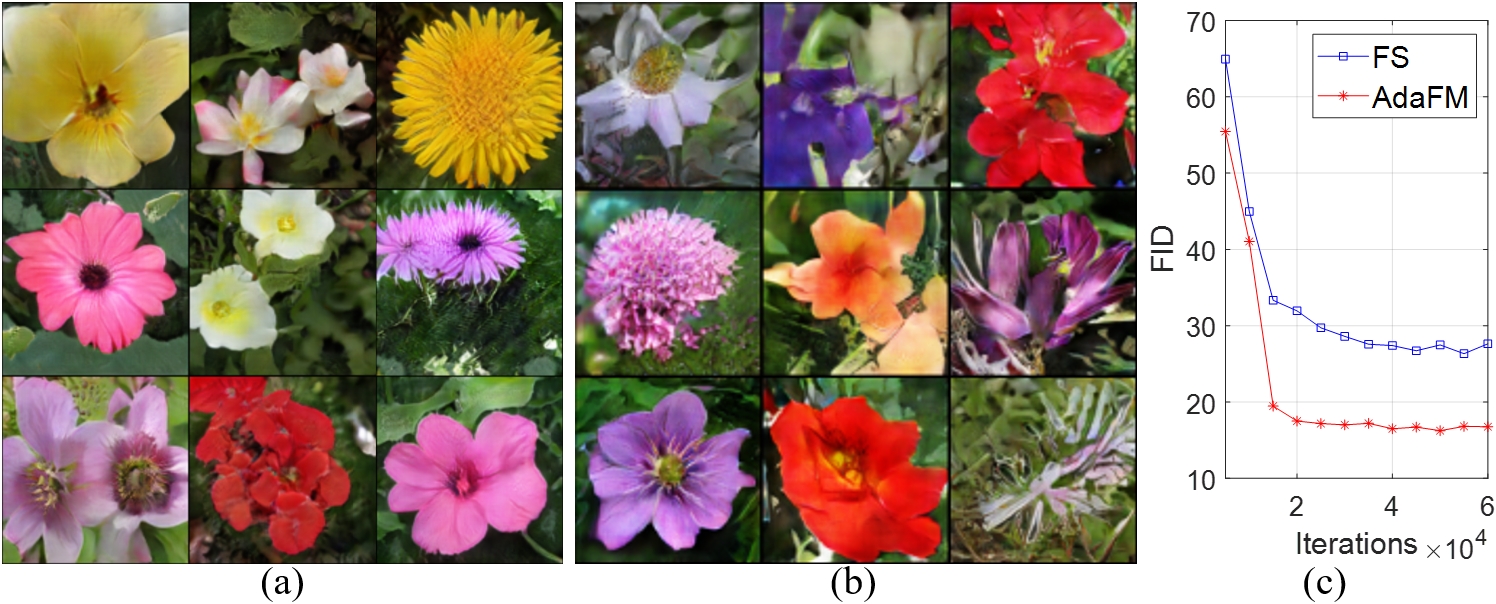}
	\vspace{-7mm}
	\caption{ 
		Generated samples from our model (a) with AdaFM, and (b) with AdaFM replaced by FS. (c) FID scores along training. 
	}
	\label{fig:Flowers_AdaCK_vs_BNC}
	\vspace{-4 mm}
\end{figure}

Another special case of AdaFM is the filter selection (FS) presented in \cite{noguchi2019image}, which employs rank-one simplification to both scale $\gammav$ and shift $\betav$.
Specifically, $\gammav = \hat \gammav \bds 1^T$ and shift $\betav = \hat \betav \bds 1^T$ with $\hat \gammav \in \Rbb^{C_{\text{out}}}$ and $\hat \betav \in \Rbb^{C_{\text{out}}}$ (see Figure \ref{fig:Network_all}(d)).
The goal of FS is to ``select'' the transferred filters $\Wten$; for example if $\hat \gammav$ is a binary vector, it literally selects among $\{\Wten_{i,:,:,:}\}$. 
As FS doesn't modulate among $C_{\text{in}}$ input channels, its basic assumption is that the source and target domains share the same correlation among those channels, which might not be true.
See the illustrative example in Figure \ref{fig:AdaFM_apple}, where the source/target domain has the basic pattern/filter of an almost-red/almost-green square (the shape is the same); it's clear simply selecting (FS) the source filter won't deliver a match to the target, appealing for a modulation (AdaFM) among (input) channels.
Figure \ref{fig:Flowers_AdaCK_vs_BNC} shows results from models with FS and AdaFM (using the same architecture in Figure \ref{fig:Network_all}(h)). It's clear that AdaFM brings boosted performance, empirically supporting the above intuition that the basic shape/pattern within each ${\Wten_{i,j,:,:}}$ are generally applicable while the correlation among $i$-,$j$-channels may be target-specific (this is further verified by AdaFM delivering boosted performance on other datasets in the experiments).

\vspace{-1mm}
\section{Experiments}

\begin{figure*}[tb]
	\centering
	\includegraphics[height=0.4\columnwidth]{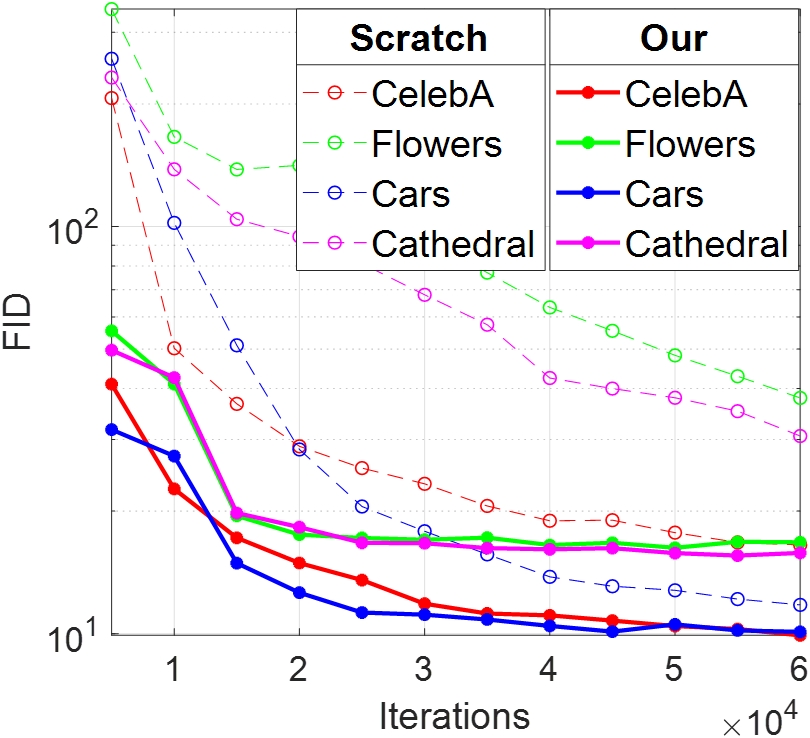}
	\qquad
	\includegraphics[height=0.4\columnwidth]{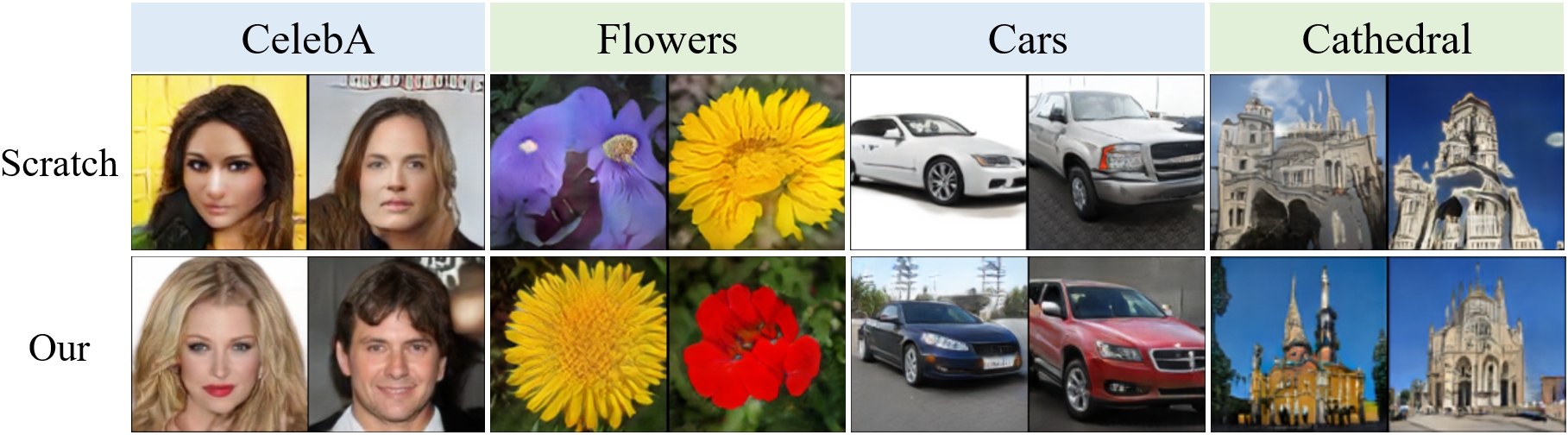}
	\vspace{-2mm}
	\caption{
		FID scores (left) and generated images (right) of Scratch and our method on 4 target datasets.
		The transferred general part from the pretrained GP-GAN model dramatically accelerates the training, leading to better performance. 
	}
	\label{fig:scratch_vs_our_big4}
	\vspace{-4 mm}
\end{figure*}

\begin{table*}[b]
	\vspace{-2mm}
	\begin{minipage}{0.57\linewidth}
		\centering
		\includegraphics[width=0.9\columnwidth]{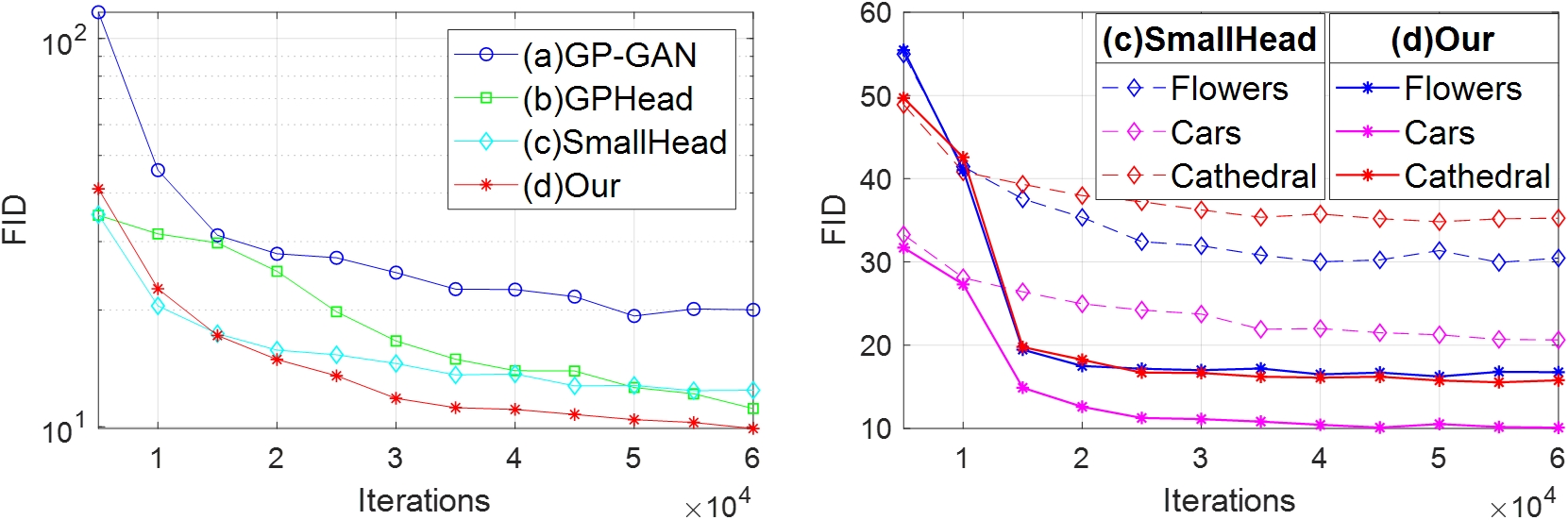}	
		\vspace{-3mm}
		\captionof{figure}{FID scores from the ablation studies of our method on CelebA (left) and the 3 small datasets: Flower, Cars, and Cathedral (right).
		}
		\label{fig:fid_our_step_by_step}
	\end{minipage}\quad
	\begin{minipage}{0.4\linewidth}
		\centering
		\setcounter{table}{1}
		\caption{FID scores from ablation studies of our method after 60,000 training iterations. 
			Lower is better. 
			\label{tab:our_step_by_step}}
		\vspace{2mm}
		\resizebox{\hsize}{!}{
			\begin{tabular}{lcccc}
				\hline\hline
				Method$\backslash$Target   & CelebA   & Flowers & Cars & Cathedral  \\
				\hline
				(a): GP-GAN   & 19.48    & failed  & failed  & failed  \\
				(b): GPHead   & 11.15    & failed  & failed  & failed  \\
				(c): SmallHead 	& 12.42    & 29.94   & 20.64   & 34.83   \\
				(d): Our        & \textbf{9.90}     & \textbf{16.76}   &  \textbf{10.10}    & \textbf{15.78}   \\
				\hline\hline 
			\end{tabular}
		}
	\end{minipage}
\end{table*}

Taking natural image generation as an illustrative example, we demonstrate the effectiveness of the proposed techniques by transferring the source GP-GAN model pretrained on the large-scale ImageNet (containing $1.2$ million images from 1,000 classes) to facilitate generation in perceptually-distinct target domains with
($i$) four smaller datasets, \ie CelebA \cite{liu2015deep} (202,599), Flowers \cite{nilsback2008automated} (8,189), Cars \cite{KrauseStarkDengFei-Fei_3DRR2013} (8,144), and Cathedral \cite{zhou2014learning} (7,350);
($ii$) their modified variants containing only 1,000 images;
and ($iii$) two extremely limited datasets consisting of 25 images (following \cite{noguchi2019image}).

The experiments proceed by
($i$) demonstrating the advantage of our method over existing approaches;
($ii$) conducting ablation studies to analyze the contribution of each component of our method;
($iii$) verifying the proposed techniques in challenging settings with only 1,000 or 25 target images;
($iv$) analyzing why/how AdaFM leads to boosted performance;
and ($v$) illustrating the potential for exploiting the tailored specific part of our model for data augmentation for applications with limited data.
Generated images and FID scores \cite{heusel2017gans} are used to evaluate the generative performance.
Detailed experimental settings and more results are provided in the Appendix.
Code is available at \url{github.com/MiaoyunZhao/GANTransferLimitedData}.

\vspace{-1mm}
\subsection{Comparisons with Existing Methods}
\label{sec:compare_with_existing_method}
\vspace{-1mm}

\begin{table}[tb]
	\setcounter{table}{0}
	\vspace{-2mm}
	\centering
	\caption{FID scores of the compared methods after 60,000 training iterations. Lower is better. 
		``Failed'' means training/mode collapse.
		\label{tab:final_fid_other}}	
	\vspace{-1 mm}
	\resizebox{0.98\hsize}{!}{
		\begin{tabular}{ccccc}
			\hline\hline
			Method$\backslash$Target	& CelebA   & Flowers  & Cars     & Cathedral    \\
			\hline
			TransferGAN   &  18.69    & failed   & failed   & failed       \\
			Scratch    & 16.51    & 29.65    & 11.77    & 30.59        \\
			Our           & \textbf{9.90}   & \textbf{16.76}    & \textbf{10.10}   & \textbf{15.78}   \\
			\hline\hline
		\end{tabular}
	}
	\vspace{-4 mm}
\end{table}

To demonstrate our contributions over existing approaches, we compare our method with 
($i$) TransferGAN \cite{wang2018transferring}, which initializes with the pretrained GP-GAN model (accordingly the same network architecture is adopted; refer to Figure \ref{fig:Network_all}(f)), followed by fine-tuning all parameters on the target data. We also consider 
 ($ii$) Scratch, which trains a model with the same architecture as ours (see Figure \ref{fig:Network_all}(h)) from scratch with the target data.

The experimental results are shown in Figure \ref{fig:scratch_vs_our_big4}, with the final FID scores summarized in Table \ref{tab:final_fid_other}.
Since TransferGAN employs the source (large) GP-GAN architecture, it may suffer from overfitting if the target data are too limited, which manifests as training/mode collapse; accordingly, TransferGAN fails on the 3 small datasets: Flowers, Cars, and Cathedral.
By comparison, thanks to the tailored specific part, both Scratch and our method train stably on all target datasets, as shown in Figure \ref{fig:scratch_vs_our_big4}.
Compared to Scratch, our method shows dramatically increased training efficiency, thanks to the transferred low-level filters, and significantly improved generative quality (much better FIDs in Table \ref{tab:final_fid_other}), which are attributed to both the transferred general part and a better adaption to target domains with AdaFM.

\vspace{-1.5mm}
\subsection{Ablation Study of Our Method}
\vspace{-1mm}

\begin{table*}[!ht]
	\begin{minipage}{0.62\linewidth}
		\centering
		\includegraphics[width=0.98\columnwidth]{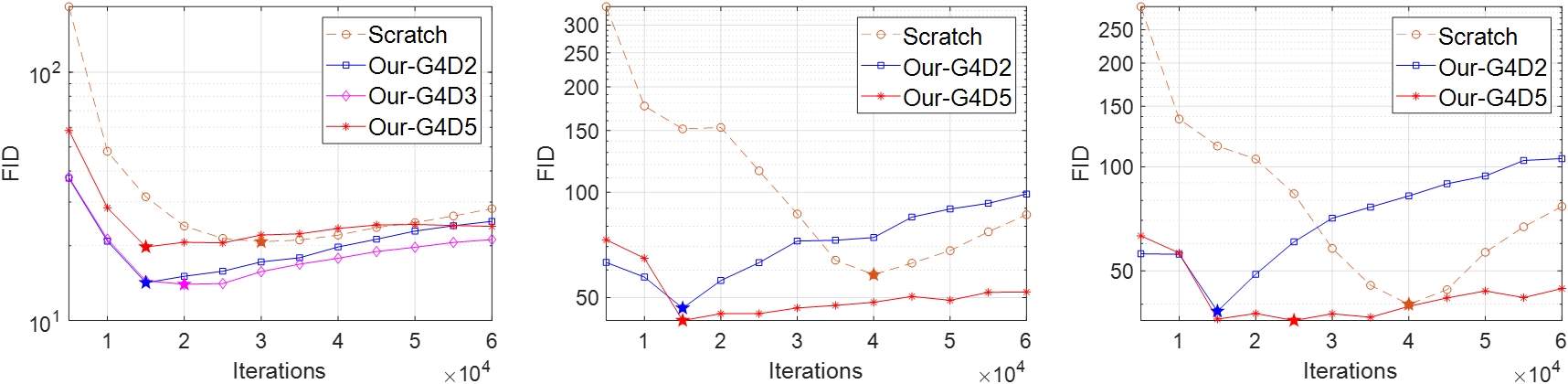}
		\vspace{-2mm}
		\captionof{figure}{ 
			FID scores on CelebA-1K (left), Flowers-1K (center), and Cathedral-1K (right).
			The best FID achieved is marked with a star.
		}
		\label{fig:FID_1K_3dataset}
	\end{minipage}
	\begin{minipage}{0.37\linewidth}
		\setcounter{table}{2}
		\caption{
			The best FID achieved within 60,000 training iterations on the limited-1K datasets.
					Lower is better. 
			\label{tab:1K_final_fids}}
		\vspace{0.2 cm}	
		\centering
		\resizebox{\hsize}{!}{
			\begin{tabular}{ccccc}
				\hline\hline
				Method$\backslash$Target & CelebA-1K   & Flowers-1K   & Cathedral-1K    \\
				\hline
				Scratch    	& 	20.75  	& 	58.18	& 	39.97       \\
				Our-G4D2   	& 	14.19   & 	46.68	& 	38.17     \\
				Our-G4D3    & 	\textbf{13.99}   & 	-    & 	-   \\
				Our-G4D5    & 	19.77   & 	\textbf{43.05}    & 	\textbf{35.88}   \\
				\hline\hline
			\end{tabular}
		}
	\end{minipage}
\vspace{-4 mm}
\end{table*}

\begin{figure*}[b]
	\vspace{-2mm}
	\centering
	\includegraphics[width=1.6\columnwidth]{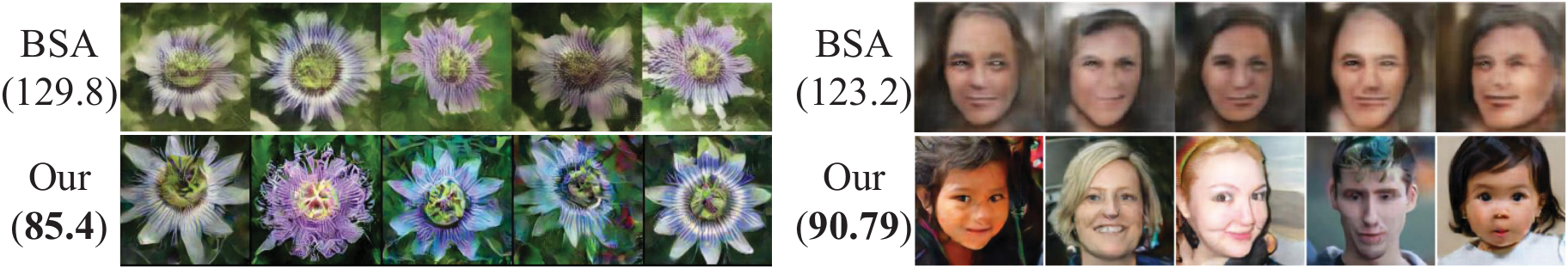}
	\vspace{-2mm}
	\caption{
		Generated images and FID scores of the compared methods on Flowers-25 (left) and FFHQ-25 (right).
		The BSA results are copied from the original paper \cite{noguchi2019image}.
	}
	\label{fig:BSA_vs_our_25images_samples}
	\vspace{-2 mm}
\end{figure*}

\begin{figure*}[b]
	\centering
	\includegraphics[width=1.6\columnwidth]{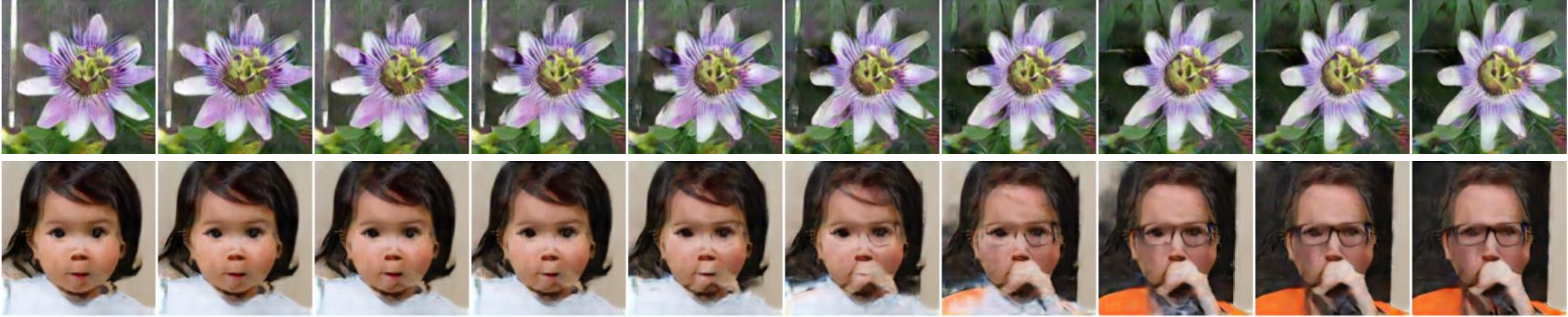}
	\vspace{-2mm}
	\caption{
		Interpolations between two random samples from our method on Flowers-25 (first row) and FFHQ-25 (second row).
	}
	\label{fig:BSA_vs_our_25images_interplations}
\end{figure*}

To reveal how each component contributes to the excellent performance of our method, we consider four experimental settings in a sequential manner. 
\textit{(a)} \textbf{GP-GAN}: adopt the GP-GAN architecture (similar to Figure \ref{fig:Network_all}(f) but all parameters are trainable and randomly initialized), used as a baseline where no low-level filters are transferred.	
\textit{(b)} \textbf{GPHead}: use the model in Figure \ref{fig:Network_all}(f), to demonstrate the contribution of the transferred general part.
\textit{(c)} \textbf{SmallHead}: employ the model in Figure \ref{fig:Network_all}(g), to reveal the contribution of the tailored specific part. 
\textit{(d)} \textbf{Our}: leverage the model in Figure \ref{fig:Network_all}(h), to show the contribution of the presented AdaFM.

The FID curves during training and the final FID scores of the compared methods are shown in Figure \ref{fig:fid_our_step_by_step} and Table \ref{tab:our_step_by_step}, respectively.
By comparing GP-GAN with GPHead on CelebA, it's clear that the transferred general part contributes by dramatically increasing the training efficiency and by delivering better generative performance; this is consistent with what's revealed in the previous section (compare Scratch with Our in Figure \ref{fig:scratch_vs_our_big4} and Table \ref{tab:final_fid_other}).
Comparing SmallHead to both GPHead and GP-GAN in Table \ref{tab:our_step_by_step} indicates that the tailored specific part helps alleviate overfitting and accordingly delivers stable training.
By better adapting the transferred general part to the target domains, the proposed AdaFM contributes most to the boosted performance (compare SmallHead with Our in Figure \ref{fig:fid_our_step_by_step} and Table \ref{tab:our_step_by_step}), empirically confirming our intuition in Section \ref{sec:Ada_FM}.

\vspace{-1mm}
\subsection{Generation with Extremely Limited Data}

To verify the effectiveness of the proposed techniques in more challenging settings, we consider generation with only 1,000 or 25 target samples. 
Specifically, we randomly select 1,000 images from CelebA, Flowers, and Cathedral to form their limited-1K variants, termed CelebA-1K, Flowers-1K, and Cathedral-1K, respectively.
Since TransferGAN fails when given about 8,000 target images (see Section \ref{sec:compare_with_existing_method}), we omit it and only compare our method with Scratch on these 1K variants.
Regarding the extremely limited setup with 25 samples, we follow \citet{noguchi2019image} to select 25 images from Flowers and FFHQ \cite{Karras_2019_CVPR} to form the Flowers-25 and FFHQ-25 datasets, on which their BSA and our method are compared.

The FID curves versus training iterations on the 1K datasets are shown in Figure \ref{fig:FID_1K_3dataset}, with the lowest FIDs summarized in Table \ref{tab:1K_final_fids}. 
In this challenging setting, both Scratch and our method with the G4D2 general part (labeled Our-G4D2) suffer from overfitting. 
Scratch suffers more due to more trainable parameters; as our method has a much higher training efficiency, a false impression may potentially arise \cite{cong2019go,mo2020freeze}; for clarity, see the significantly improved best performance of our method and the more gentle ending slope of its FID curves.
To alleviate overfitting, we transfer more discriminator filters from the pretrained GP-GAN model, with the results also given in Figure \ref{fig:FID_1K_3dataset} and Table \ref{tab:1K_final_fids}. It's clear that intuitive patterns emerge, \ie less data appeal for more transferred information. 
On the other hand, the comparable FIDs of Our-G4D2 (see Table \ref{tab:1K_final_fids}) indicates that the G4D2 general part discovered in Section \ref{sec:reuse_Big} works fairly well on these 1K datasets.
Concerning early stopping for generation with limited data, we empirically find that the discriminator loss may be leveraged for that goal in our setup (see Appendix \ref{secapp:early_stop_limitedata} for detailed discussions).

On Flowers-25 and FFHQ-25, since the target data are extremely limited in quantity, we transfer more filters (\ie G4D6) from the pretrained GP-GAN model and apply GP (gradient penalty) on both real and fake samples to alleviate overfitting (see Appendix \ref{appsec:25sample} for detailed settings).
Figure \ref{fig:BSA_vs_our_25images_samples} shows the generated samples and FID scores from the BSA \cite{noguchi2019image} and our method.
It's clear that our method with the G4D6 general part works reasonably well even in such settings with extremely limited data, with a much better performance than the L1/Perceptual-loss-based BSA. 
To illustrate the learned data manifold, Figure \ref{fig:BSA_vs_our_25images_interplations} shows the smooth interpolations between two random samples from our method, demonstrating the effectiveness of the proposed techniques on generation with extremely limited data.

\vspace{-1mm}
\subsection{Analysis of AdaFM and Style Augmentation with the Tailored Specific Part}
\label{sec:EXP_AdaFM_style}

\begin{figure}[t]
	\centering
		\includegraphics[width=\columnwidth]{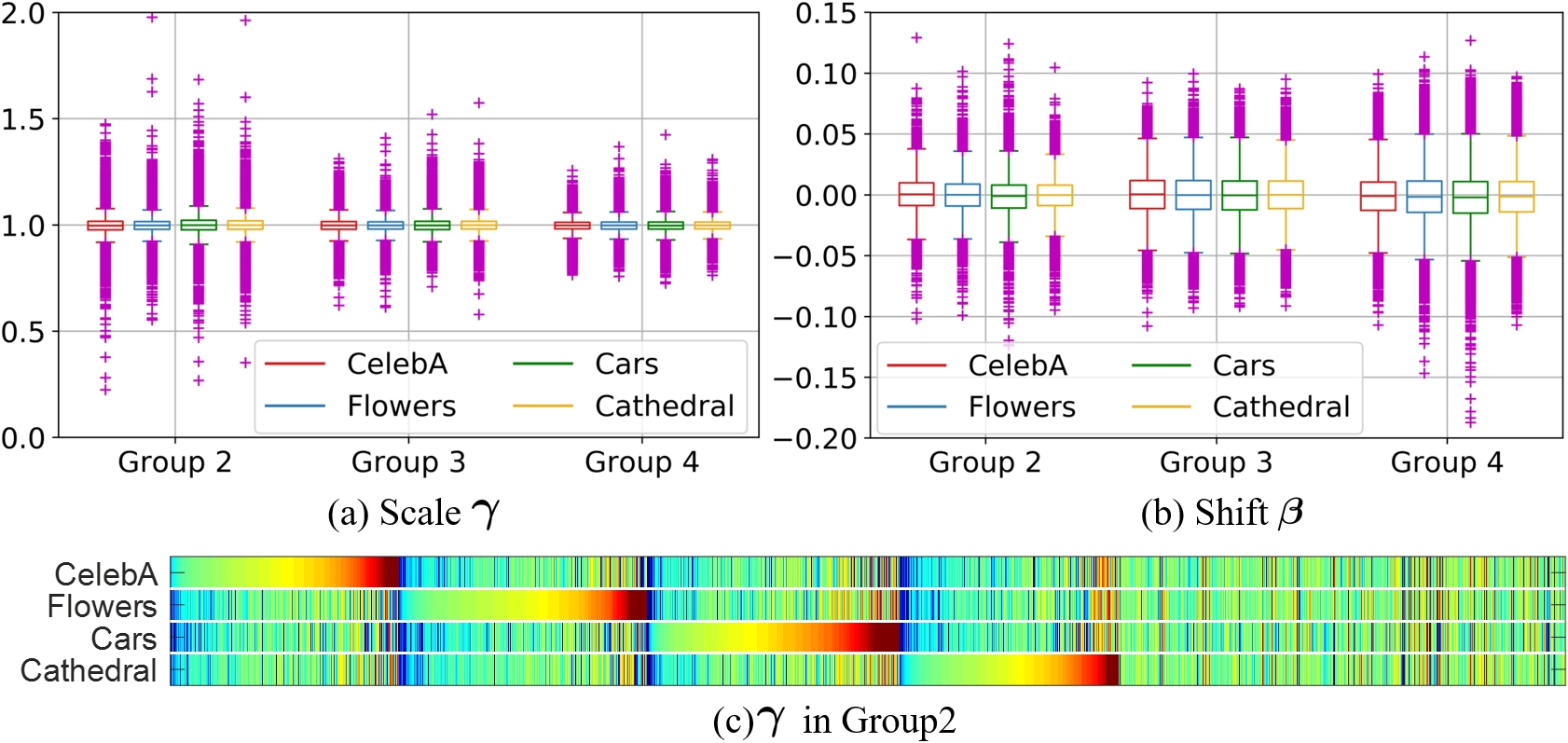}
		\vspace{-6mm}
		\caption{ 
			Boxplots of the learned scale $\gammav$ (a) and shift $\betav$ (b) within each group of the generator general part. (c) Sorted comparison of the $\gammav$ learned on different datasets (see Appendix \ref{gamma_matrix} for details).
		}
		\label{fig:AdaKC_distribute}
	\vspace{-6 mm}
\end{figure}

To better understand why adopting AdaFM in the transferred general part of our model leads to boosted performance, we summarize in Figures \ref{fig:AdaKC_distribute}(a) and \ref{fig:AdaKC_distribute}(b) the learned scale $\gammav$ and shift $\betav$ from different groups of the generator general part.
Apparently, all transferred filters are used in the target domains (no zero-valued $\gammav$) but with modulations ($\gammav/\betav$ has values around $1/0$).
As AdaFM delivers boosted performance, it's clear those modulations are crucial for a successful transfer from source to target domain, confirming our intuition in Section \ref{sec:Ada_FM}.
To illustrate how $\gammav/\betav$ behaves on different target datasets, we show in Figure \ref{fig:AdaKC_distribute}(c) the sorted comparisons of the learned $\gammav$ in Group 2; as expected, different datasets prefer different modulations, justifying the necessity of AdaFM and its performance gain.
Concerning further demonstration of AdaFM and medical/biological applications with gray-scale images, we conduct another experiment on a gray-scale variant of Cathedral (results are given in Appendix \ref{secapp:black_white} due to space constraints); we find that without AdaFM to adapt the transferred filters, worse (blurry and messy) details are observed in the generated images (refer also to Figure \ref{fig:Flowers_AdaCK_vs_BNC}), likely because of the mismatched correlation among channels between source and target domains.

\begin{figure}[t]
	\centering
		\includegraphics[width=0.98\columnwidth]{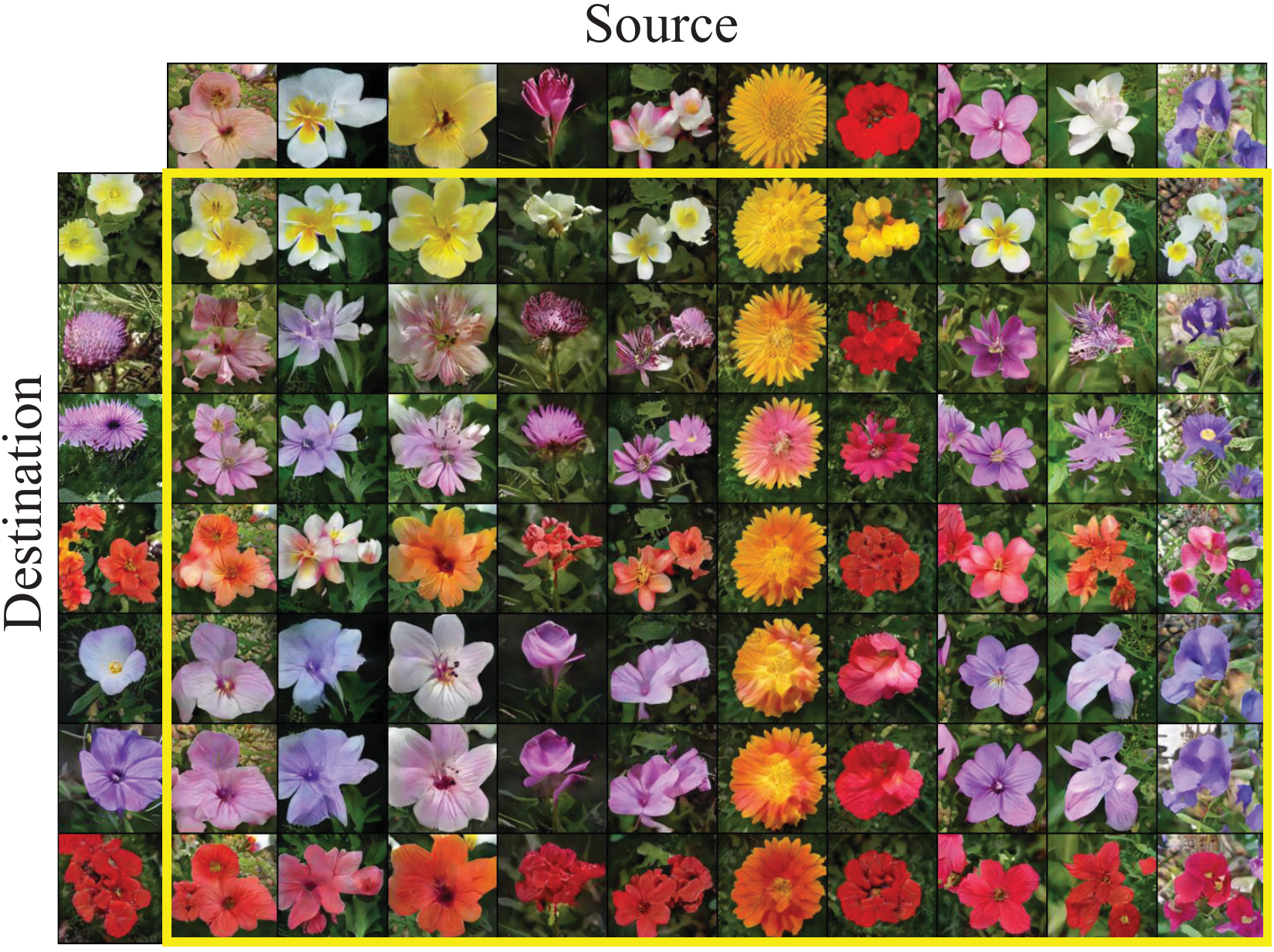}
		\vspace{-2 mm}
		\caption{Style mixing on Flowers via the tailored specific part of our model.
			The ``Source'' sample controls flower shape, location, and background, while the ``Destination'' sample controls color and petal details.
		}
		\label{fig:mix_style_Flowers}
 	\vspace{-6 mm}
\end{figure}

To reveal the potential in exploiting the tailored specific part of our model for data augmentation for applications with limited data, we conduct style mixing with the specific part, following \cite{Karras_2019_CVPR}. 
Figure \ref{fig:mix_style_Flowers} shows the results on Flowers (see Appendix \ref{sec:style_mixing} for details and more results). 
It's clear that style mixing enables synthesizing a vast set of new images via style/attribute combinations.
Therefore, the tailored specific part of our model can be exploited for a diverse synthetic augmentation, which is believed extremely appealing for downstream  applications with limited data.

\vspace{-2mm}
\section{Conclusions}
\vspace{-1mm}

We reveal that the valuable information (\ie the low-level filters) within GAN models pretrained on large-scale source datasets (\eg ImageNet) can be transferred to facilitate generation in perceptually-distinct target domains with limited data; this transfer is performed on both the generator and discriminator.
To alleviate overfitting due to the limited target-domain data, we employ a small specific network atop the transferred low-level filters, which enables style mixing for a diverse synthetic augmentation.
To better adapt the transferred filters to target domains, we present adaptive filter modulation (AdaFM), that delivers boosted performance on generation with limited data.
The proposed techniques are shown to work well in challenging settings with extremely limited data (\eg 1,000 or 25 samples).

%
%

\vspace{-2mm}
\section*{Acknowledgements}
\vspace{-1mm}

We thank the anonymous reviewers for their constructive comments. 
The research was supported by part by DARPA, DOE, NIH, NSF and ONR. The Titan Xp GPU used was donated by the NVIDIA Corporation.

\bibliography{ReferencesCong}
\bibliographystyle{icml2020}

\clearpage
\newpage

\appendix

\twocolumn[
\icmltitle{Appendix of \\
	On Leveraging Pretrained GANs for Generation with Limited Data
}

\centering{
	
	\textbf{Miaoyun Zhao, Yulai Cong, Lawrence Carin
		\\Department of ECE, Duke University}
	
}
\vskip 0.3in
]


\section{Experimental Settings}
\label{secapp:exp_set}

For all experiments if not specified, we inherit the experimental settings of GP-GAN \cite{mescheder2018training}. 
All training images are resized to $128\times128$ for consistency. For the discriminator, gradient penalty on real samples ($R_1$-regularizer) is adopted with  the regularization parameter $\gamma=10.0$.
We use the Adam optimizer with learning rate $1\times10^{-4}$ and coefficients $(\beta_1,\beta_2)=(0.0,0.99)$. 
$60,000$ iterations are used for training. 
Because of limited computation power, we use a batch size of $16$. 
For the CelebA dataset, the dimension of the latent vector $\zv$ is set to $256$; while for the small datasets (\ie Flowers, Cars, and Cathedral) and their 1K variants, that dimension is set to $64$.

The experimental settings for the extremely limited datasets, \ie Flowers-25 and FFHQ-25, are provided in Appendix \ref{appsec:25sample}.

The FID scores are calculated based on $10,000/8,189/8,144/7,350$ real and generated images on CelebA, Flowers, Cars, and Cathedral, respectively. 
The same FID calculations are employed for experiments on the corresponding 1K variants.

For Scratch, we employ the same architecture as our method without the $\gammav/\betav$ AdaFM parameters, because $\gammav$s and $\betav$s are now redundant if we train the source filter $\Wten$ (refer to \eqref{eq:AdaFM} of the main manuscript).

Regarding the training of our method, we fix the scale $\gammav=1$ and shift $\betav=0$ in the first 10,000 iterations and only update the tailored specific part for a stable initialization; after that, we jointly train both the $\gammav/\betav$ AdaFM parameters and the specific part to deliver the presented results.

\subsection{On Specifying the General Part of the Discriminator for Transfer}
\label{sec:initialize}

\begin{figure}[tb]
	\begin{center}
		\includegraphics[width=0.6\columnwidth]{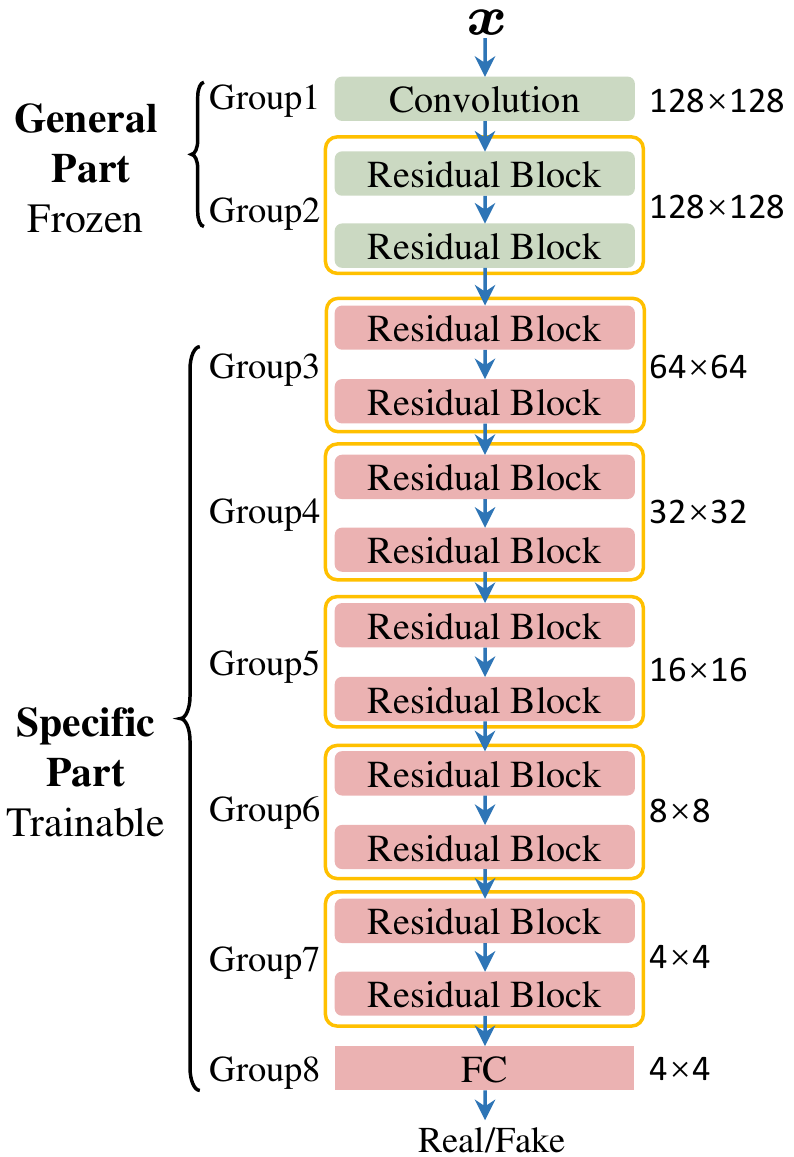}
		\caption{
			The architecture for the discriminator with the D2 general part. 
			The overall architecture is inherited from the GP-GAN.  
		}
		\label{fig:Framework_D}
	\end{center}
\end{figure}

To figure out the suitable general part of the discriminator to be transferred from the pretrained GP-GAN model to the target CelebA dataset, we design a series of experiments with increasing number of lower groups included/frozen in the transferred/frozen general part; the remaining high-level specific part is reinitialized and trained with CelebA.
The architecture for the discriminator with the D2 general part is shown in Figure \ref{fig:Framework_D}, as an illustrative example.

Based on the transferred general part of the generator and discriminator, we next reinitialize and train the remaining specific part on CelebA. The employed reinitialization are as follows.
\begin{itemize}
	\item For all layers except FC in the generator/discriminator, we use the corresponding parameters from the pretrained GP-GAN model as initialization.
	\item Regarding FC layers in generator/discriminator, since the pretrained GP-GAN model on ImageNet used a conditional-generation architecture (\ie the input of the generator FC layer consists both the noise $\zv$ and the label embedding $\yv$, whereas the discriminator FC layer has multiple heads (each corresponds to one class)), we can not directly transfer the FC parameters therein to initialize our model (without labels). Consequently, we randomly initialize both FC layers in the generator and discriminator.
\end{itemize}

\section{On Generation with Extremely Limited Data with 25 Samples}
\label{appsec:25sample}

\begin{figure}[tb]
	\begin{center}
		\includegraphics[width=0.8\columnwidth]{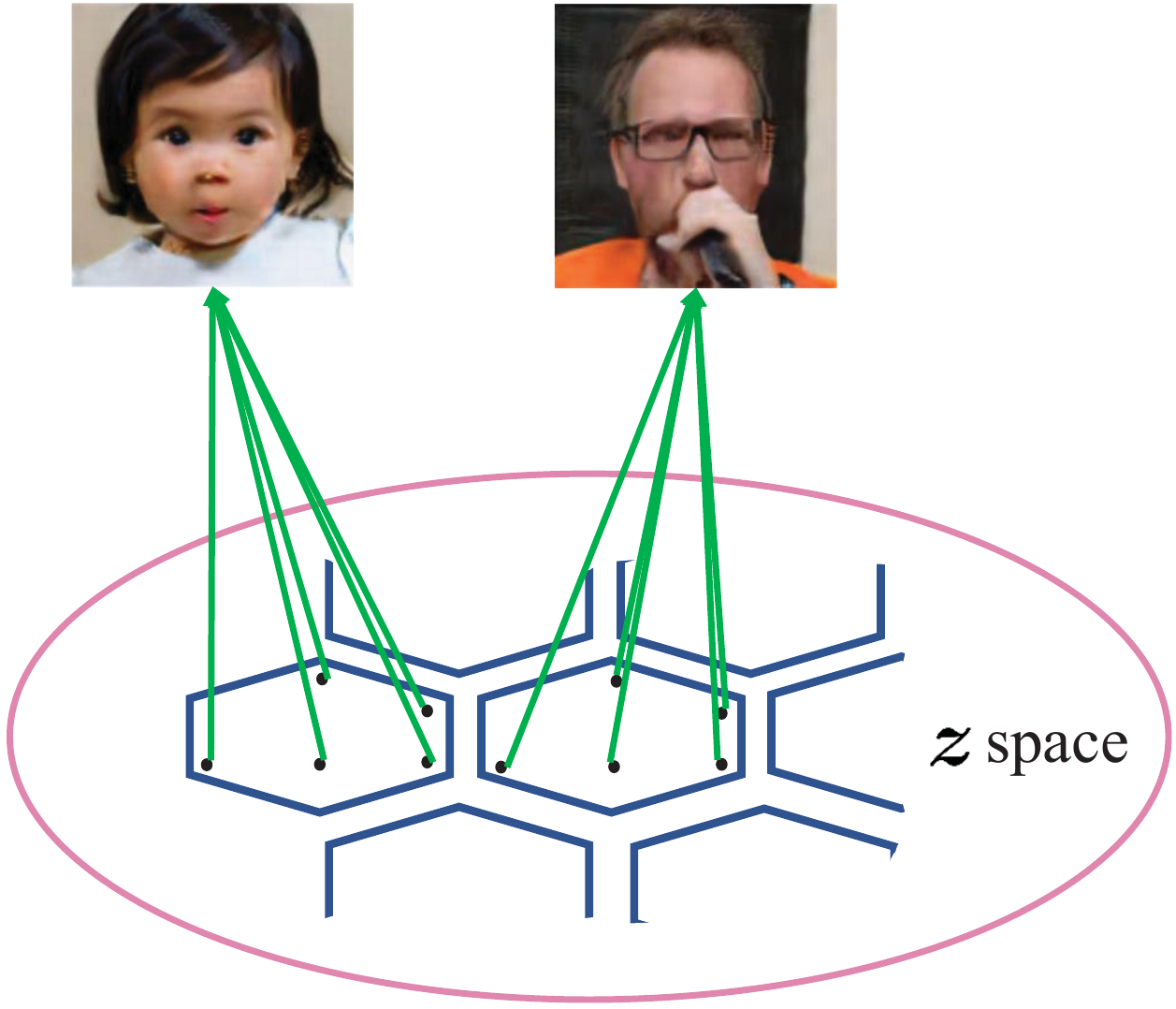}
		\caption{
			Demonstration of the generative process learned on extremely limited data. 
			Since the continuous latent $\zv$-space is likely redundant, the generator often maps close latent vectors to similar outputs.  
		}
		\label{appfig:interplation_process_1}
	\end{center}
\end{figure}

\begin{figure*}[bt]
	\begin{center}
		\includegraphics[width=1.7\columnwidth]{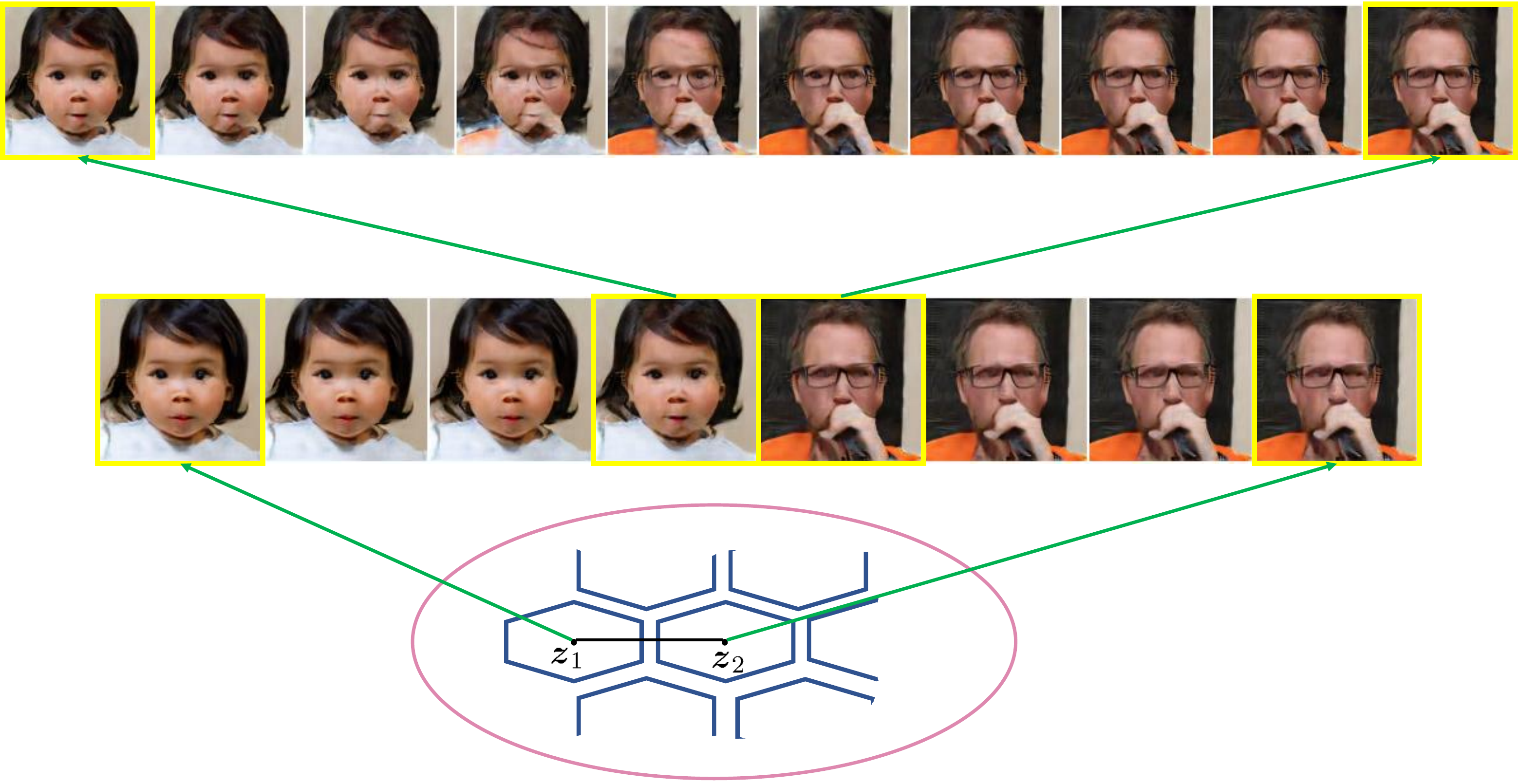}
		\caption{
			The amplification process employed to yield the smooth interpolations for our method.  
		}
		\label{appfig:interplation_process}
	\end{center}
\end{figure*}


Considering the challenging settings with extremely limited data in quantity (\ie $25$ data samples), we transfer the G4D6 general part from the pretrained GP-GAN model (termed Our-G4D6) and apply GP (gradient penalty) on both real and fake samples with the regularization parameter $\gamma = 20$. The dimension of the latent vector $\zv$ is set to $4$. 60,000 training iterations are used.

The FID scores are calculated following \cite{noguchi2019image}. For Flowers-25, 251 real passion images and 251 generated images are used to calculate the FID; for FFHQ-25, 10,000 real face images and 10,000 generated images are used.

Since the target data (25 samples) are extremely limited, we find that the generator managed to learn a generative mapping that captures the generation over the 25 training samples, as illustrated in Figure \ref{appfig:interplation_process_1}.
As the latent $\zv$-space is continuous and likely redundant for the extremely limited data, the learned generator often maps close latent vectors to a similar output.  
Regarding the interpolations shown in Figure \ref{fig:BSA_vs_our_25images_interplations} of the main paper, we use an amplification process (see Figure \ref{appfig:interplation_process}) to get the presented results. Note Figure \ref{appfig:interplation_process} also empirically verifies the above intuition.
The demonstrated results are as expected, because, on one hand, only 25 training images are available, while on the other hand, the gradient penalty applied to discriminator (in addition to regularizations from the proposed techniques) implicitly imposes smoothness to the output space of the generator.

\section{More Analysis and Discussions}

\subsection{On the Worse FID of G2D0 than That of G4D0}
\label{sec:worse_FID}

\begin{figure}[H]
	\begin{center}
		\includegraphics[width=0.8\columnwidth]{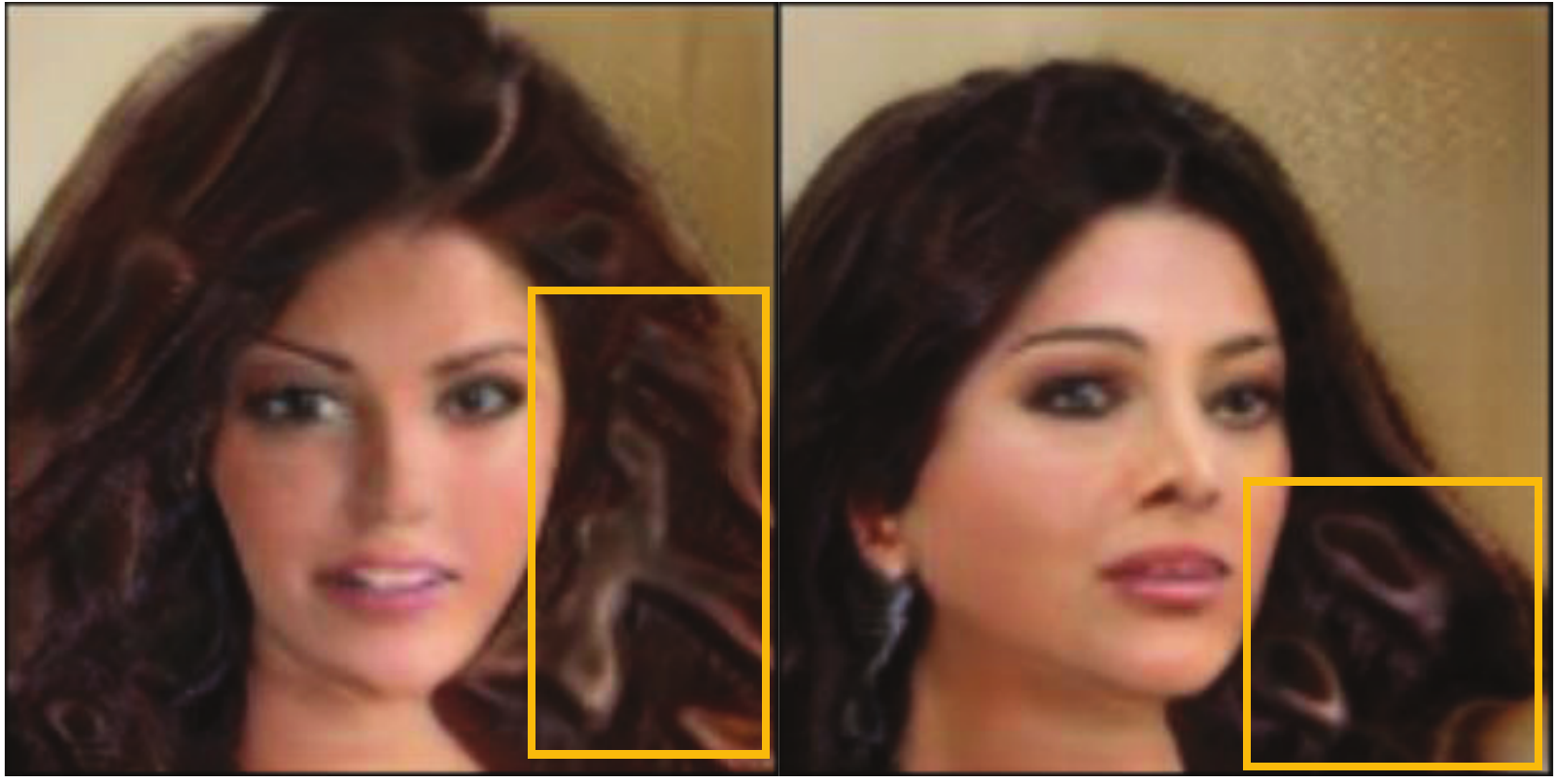}
		\caption{
			Samples generated from a model with the G2D0 general part. Water-spot shaped textures appear in the hair area (see the yellow boxes).   
		}
		\label{fig:water}
	\end{center}
\end{figure}

The worse FID of G2D0 is believed caused by the insufficiently trained low-level filters, which are time-consuming and data-demanding to train. Specifically, by taking a close look at the generated samples, we find that
\begin{itemize}
	\item[-] there are generated samples that look similar to each other, indicating a relatively low generative diversity;
	\item[-] most of the generated samples contain strange textures that look like water spots, as shown in Figure \ref{fig:water}.
\end{itemize}
Such phenomena are deemed to have negative effects on the FID score.

\begin{figure*}[t]
	\begin{center}
		\includegraphics[width=1.9\columnwidth]{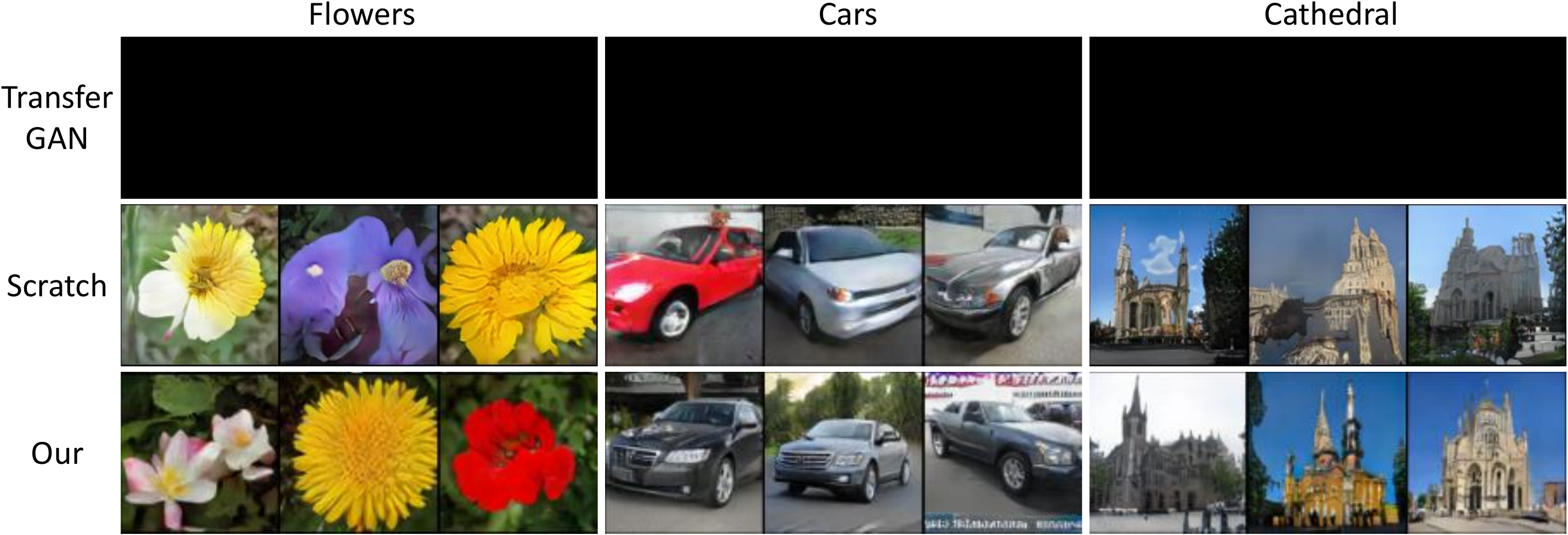}
		\caption{
			TransferGAN shows mode/training collapse on the three small datasets.   
		}
		\label{fig:transferGAN_fail}
	\end{center}
		\vspace{-5 mm}
\end{figure*}

\subsection{On Selecting the Optimal GmDn with the Best Generalization}
\label{sec:select_GmDn}

\begin{figure}[H]
	\begin{center}
		\includegraphics[width=0.8\columnwidth]{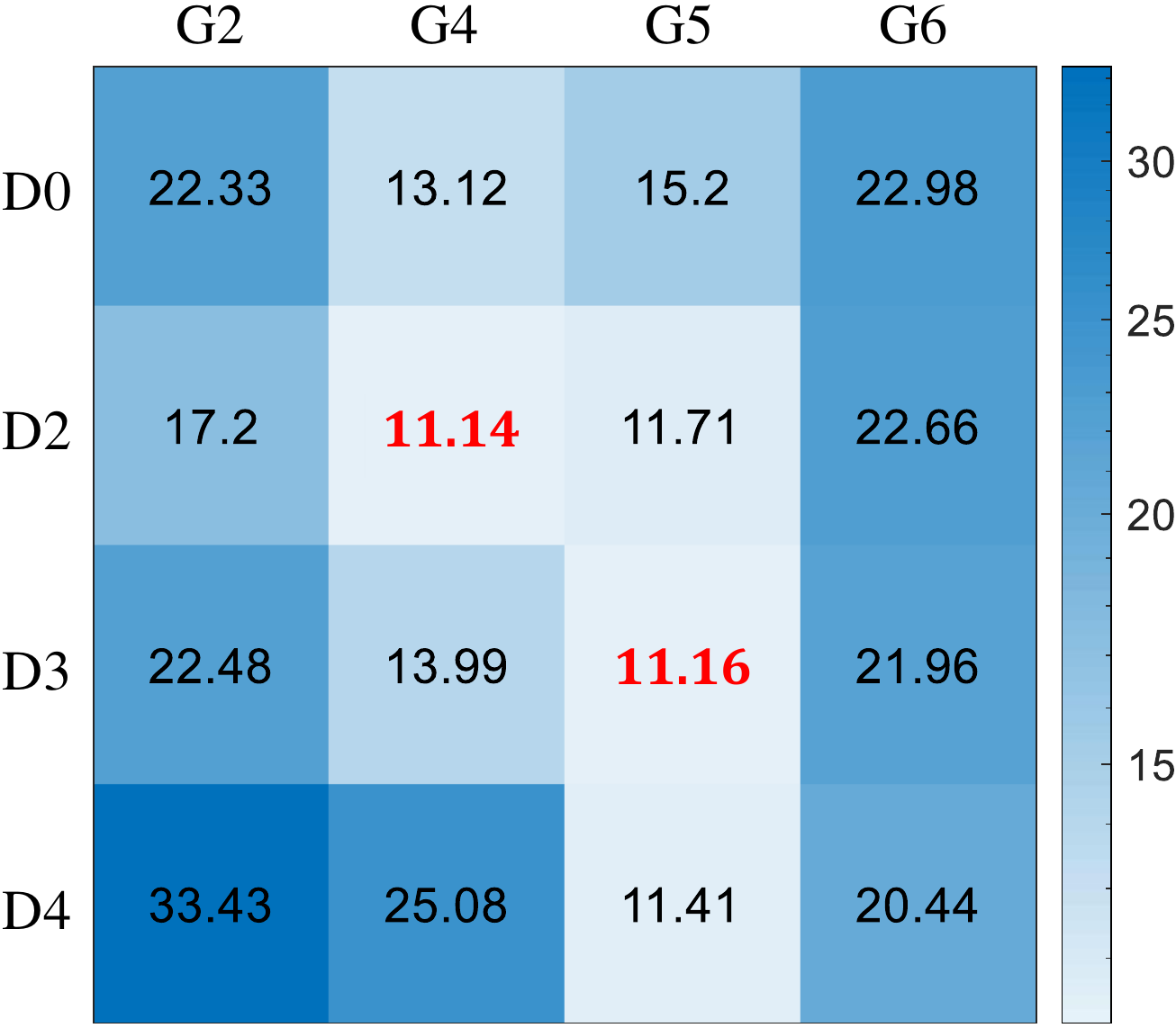}
		\caption{
			FID scores from various settings for GmDn. 
			GmDn indicates freezing the lower $m/n$ groups as the general part of generator/discriminator.
			It's clear that the adopted G4D2 general part is a reasonable choice for transfer.
		}
		\label{fig:GmDn_FIDs}
	\end{center}
\end{figure}

Supplementing Figures \ref{fig:fix_G_results} and \ref{fig:fix_D_results} of the main paper, we evaluate various settings of GmDn for the transfered general part, with the same experimental settings of Section \ref{sec:proposed_method} of the main paper. 
The corresponding FID scores are summarized in Figure \ref{fig:GmDn_FIDs}, where one may observe interesting patterns of the transfer properties of the pretrained GAN model. 
\begin{itemize}
	\item It seems that the performance is, in general, more sensitive to the setting of Gm than that of Dn, meaning that the generator general part may play a more important role in generation tasks. 
	\item Besides, it's clear that compromise arises in both Gm and Dn directions; this is expected as the low-level filters are more generally applicable while the high-level ones are more domain specific.
	\item Moreover, it seems interesting correlations exist between Gm (generator general part) and Dn (discriminator general part), which might be worthy of future explorations.
	\item Finally, the G4D2 general part delivers the best performance among the tested settings, justifying its adoption in the main paper.
\end{itemize}

It's extremely challenging (if not impossible) to choose the optimal GmDn general part that generalizes well to various target domains (beyond CelebA). 
To alleviate this concern, we'd like to point out that our AdaFM may greatly relax the requirement of an optimal GmDn, as reflected by our-G4D2's boosted performance on various target datasets. 
Empirically, we find that the G4 generator general part works reasonably well for most situations; the setting for the discriminator general part (\ie Dn) is more data-size related, \eg the D2 setting may be suitable for a target domain with $\ge$7K data samples, D5 for $\approx$1K samples, and D6 for $\approx$25 samples.

\subsection{On Figure \ref{fig:scratch_vs_our_big4} of the Main Manuscript}

Regarding the FID curves shown in Figure \ref{fig:scratch_vs_our_big4} of the main manuscript, one may concern the final performance of Scratch if it's trained for long time. To address that concern, we run Scratch on Flowers for 150,000 iterations and find that Scratch converges to a worse FID=18.1 than our method at 60,000 iterations.
Similar phenomena also manifest in Figure \ref{fig:FID_1K_3dataset} of the main manuscript.
These results empirically prove that the transferred/frozen general part deliver both improved training efficiency and better final performance.

\begin{figure*}[h]
	\begin{center}
		\includegraphics[width=1.9\columnwidth]{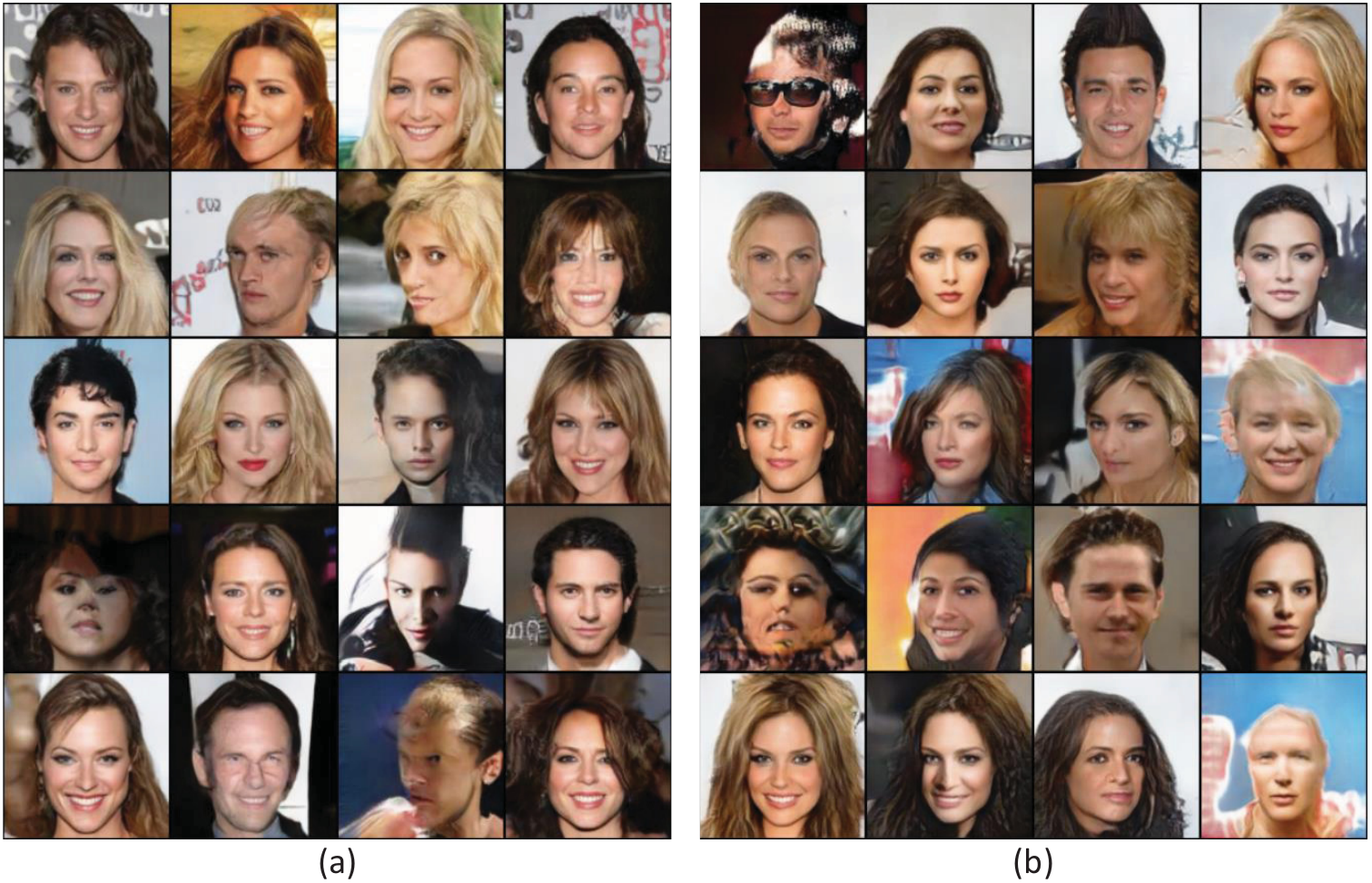}
		\caption{
			More generated samples on CelebA, supplementing Figure \ref{fig:scratch_vs_our_big4}  of the main manuscript. (a) Our; (b) Scratch. 
		}
		\label{fig:CELEBA_our_scratch}
	\end{center}
\end{figure*}
\begin{figure*}[h]
	\begin{center}
		\includegraphics[width=1.9\columnwidth]{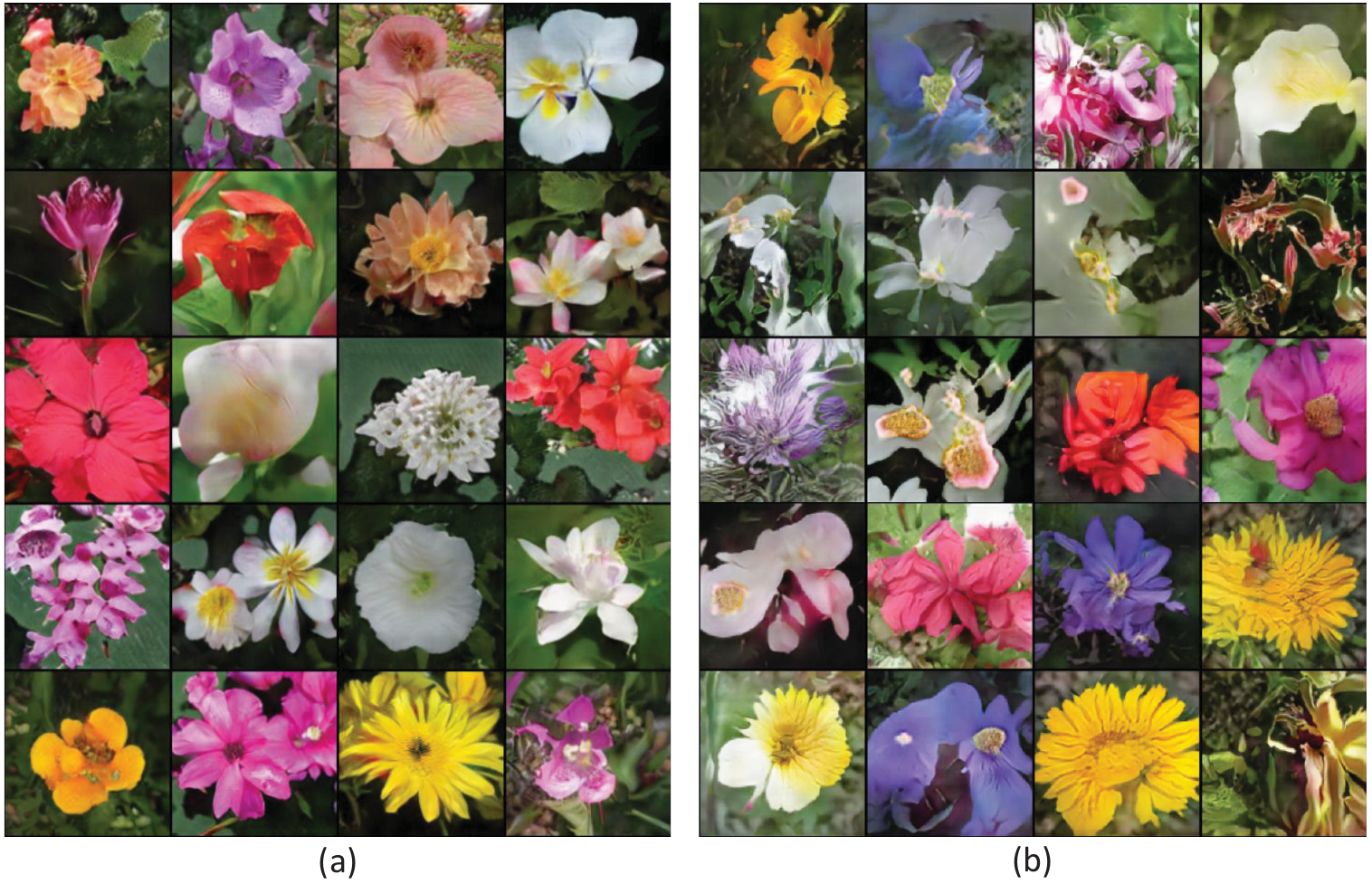}
		\caption{
			More generated samples on Flowers, supplementing Figure \ref{fig:scratch_vs_our_big4} of the main manuscript. (a) Our; (b) Scratch. 
		}
		\label{fig:Flower_our_scratch}
	\end{center}
\end{figure*}
\begin{figure*}[h]
	\begin{center}
		\includegraphics[width=1.9\columnwidth]{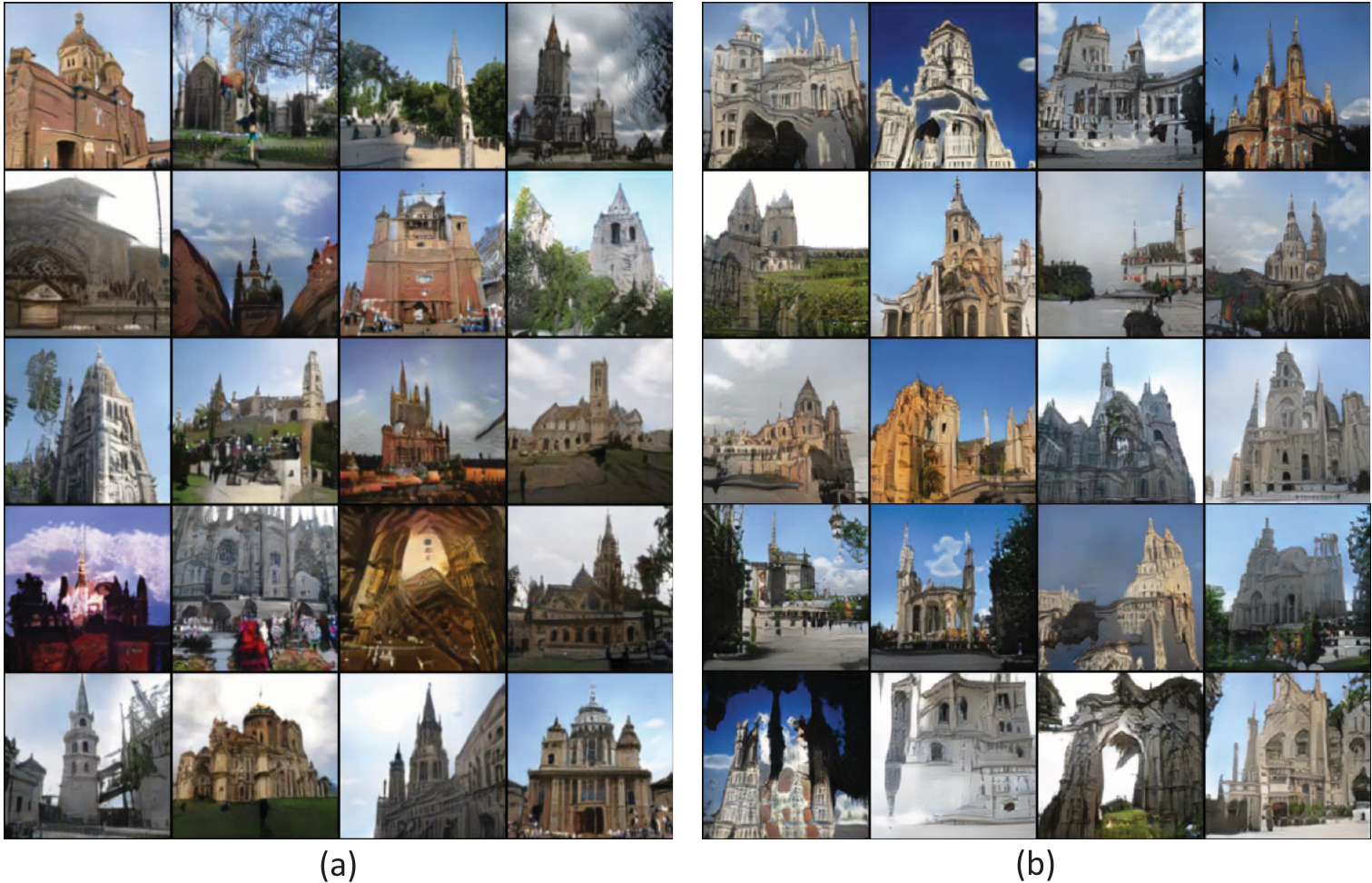}
		\caption{
			More generated samples on Cathedral, supplementing Figure \ref{fig:scratch_vs_our_big4} of the main manuscript. (a) Our; (b) Scratch. 
		}
		\label{fig:Cathedral_our_scratch}
	\end{center}
\end{figure*}

Regarding more generated samples for comparing our method with Scratch,
we show in Figure \ref{fig:CELEBA_our_scratch}, \ref{fig:Flower_our_scratch} and \ref{fig:Cathedral_our_scratch}
more randomly generated samples from our method and Scratch as a supplementary for Figure \ref{fig:scratch_vs_our_big4} of the main manuscript. 
Thanks to the transferred low-level filters and the better adaption to target domain via AdaFM, our method shows a much higher generation quality than Scratch.

\subsection{On the Failing of TransferGAN on the Three Small Datasets}
\label{sec:GP_collapse}

Since the FC layers from the pretrained GP-GAN model on ImageNet is not directly applicable to the target domain (as discussed in Section \ref{sec:initialize}), we implement the TransferGAN method by initializing parameters of all layers (except the FC layer, whose parameters are randomly initialized) with the corresponding parameters from the pretrained GP-GAN model.
A similar architecture of the pretrained GP-GAN model is therefore employed for TransferGAN.

When only a small (or limited) amount of training data are available, (\eg on the three small datasets: Flowers, Cars, and Cathedral), TransferGAN is prone to overfitting because of its large amount of trainable parameters. Specifically, the number of the trainable parameters within the TransferGAN generator is 96.1M; by comparison, the generator of our method only contains 24.4M trainable parameters.
Accordingly, TransferGAN suffers from mode/training collapse on the three small datasets, as shown in Figure \ref{fig:transferGAN_fail}.

\subsection{Early Stopping for Generation with Limited Data}
\label{secapp:early_stop_limitedata}

Concerning early stopping for generation with limited data, we find that the discriminator loss may be leveraged for that goal, as overfitting empirically manifests as a decreasing discriminator loss in our setup.

Specifically, with the GP-GAN settings, we empirically find that the discriminator loss stables roughly within $[0.8, 1.3]$ when the training is successful without clear overfitting. However, if the discriminator loss falls into $[0.5,0.7]$ and remains there for a period, overfitting likely starts arising; accordingly, that may be a proper time for early stopping.

\subsection{On Figure \ref{fig:AdaKC_distribute}(c) of the Main Manuscript}
\label{gamma_matrix}

Figure \ref{fig:AdaKC_distribute}(c) shows the sorted demonstration of the learned $\gammav$ from the last convolutional layer in Group2. We sort the $\gammav$'s learned on different target datasets as follows.
\begin{enumerate}
	\item Reshape each $\gammav$ matrix into a vector; stack these vectors into a matrix $\Mmat$, so that each row represents the $\gammav$ from a specific target dataset;
	\item Clip all the values of $\Mmat$ to $[0.9, 1.1]$ and then re-scale to $[0, 1]$ for better contrast;
	\item For the $i$-th row/target-dataset, find the set of column indexes 
	$\sv'_{i} = \{j|\forall k\ne i,\Mmat_{i,j} -\Mmat_{k,j}>0.03\}$; sort $\sv'_{i}$ according to the values $\Mmat_{i,\sv_{i}}$ to yield $\sv_{i}$;
	\item Concatenate $\{\sv_{i}\}$ with the remaining column indexes to yield the sorted indexes $\tv$; sort the columns of the matrix $\Mmat$ according to $\tv$ to deliver the presented matrix in Figure \ref{fig:AdaKC_distribute}(c).
\end{enumerate}

\subsection{Comparison of AdaFM and Weight Demodulation}
\label{appsec:compare_WD}

Supplementing Section \ref{sec:Ada_FM} of the main manuscript, we further compare the weight demodulation \cite{karras2019analyzing} with our AdaFM below.

Recall that AdaFM uses learnable matrix parameters $\{\gammav, \betav\}$ to modulate/adapt a transfered/frozen convolutional filter (see \eqref{eq:AdaFM} of the main manuscript); by comparison, the weight demodulation uses $\betav=0$ and rank-one $\gammav=\etav \sv^{T}$, where $\sv$ is parametrized as a neural network to control style and $\etav$ calculated based on $\sv$ and the convolutional filter $\Wten$ (see \eqref{eq:eta_WD} of the main manuscript). 
Therefore, a direct comparison of AdaFM and the weight demodulation is not feasible.

On one hand, it's interesting to consider generalizing our AdaFM with neural-network-parameterized $\{\gammav, \betav\}$ for better adaptation of the transfered general part, for introduction of conditional information, or for better generation like in StyleGAN2; we leave that as future research.
On the other hand, if we degrade the weight demodulation by setting $\sv$ as a learnable vector parameter, the weight demodulation may work similar to FS (see the comparison between FS and AdaFM in Figure \ref{fig:Flowers_AdaCK_vs_BNC} of the main manuscript), because both of them have almost the same flexibility.

\section{Medical/Biological Applications with Gray-Scale Images}
\label{secapp:black_white}

\begin{figure*}[h]
	\begin{center}
		\subfigure[Our method with AdaFM]{
			\includegraphics[width=1.3\columnwidth]{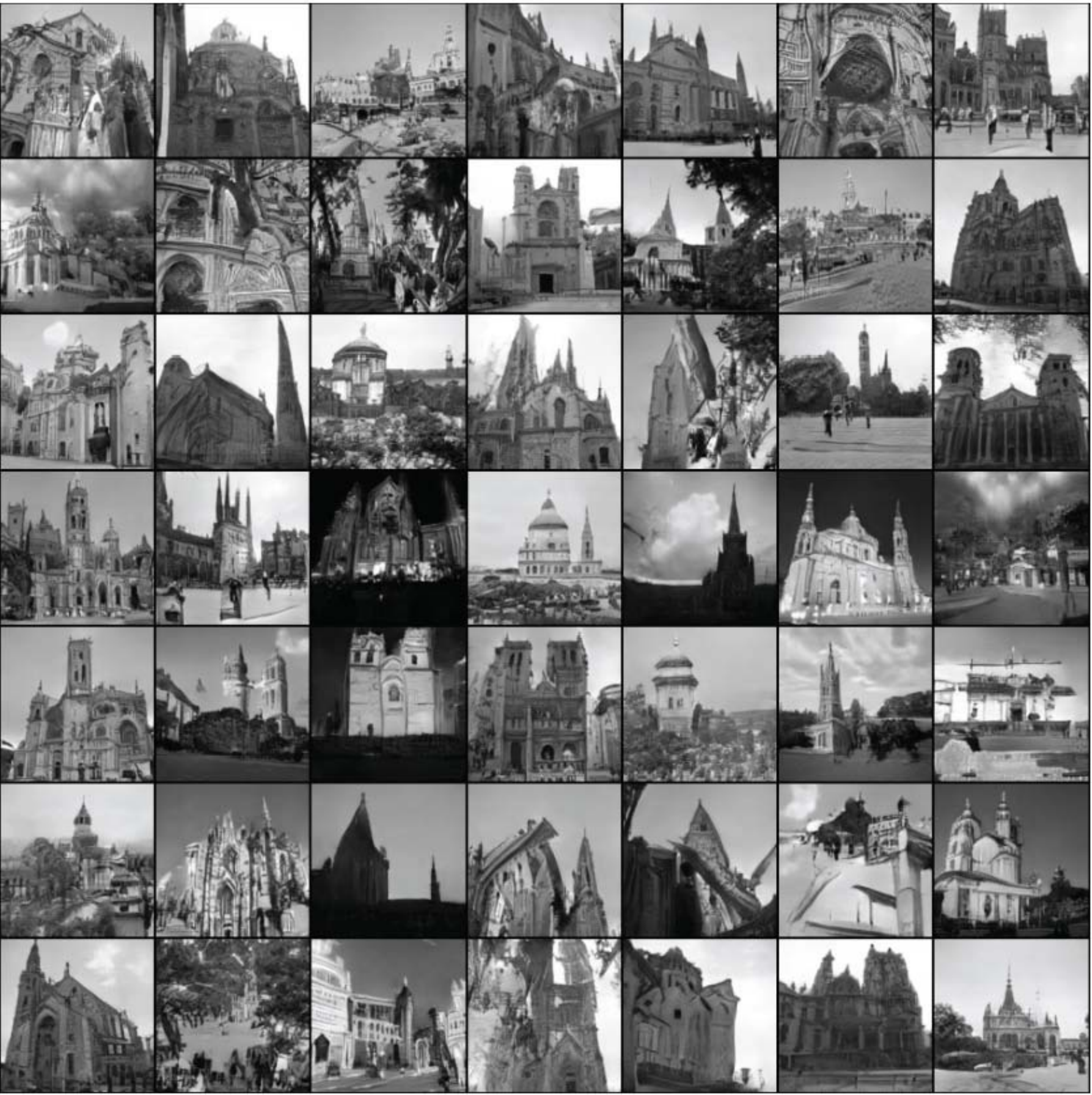}
		}
		\subfigure[SmallHead without AdaFM]{
			\includegraphics[width=1.3\columnwidth]{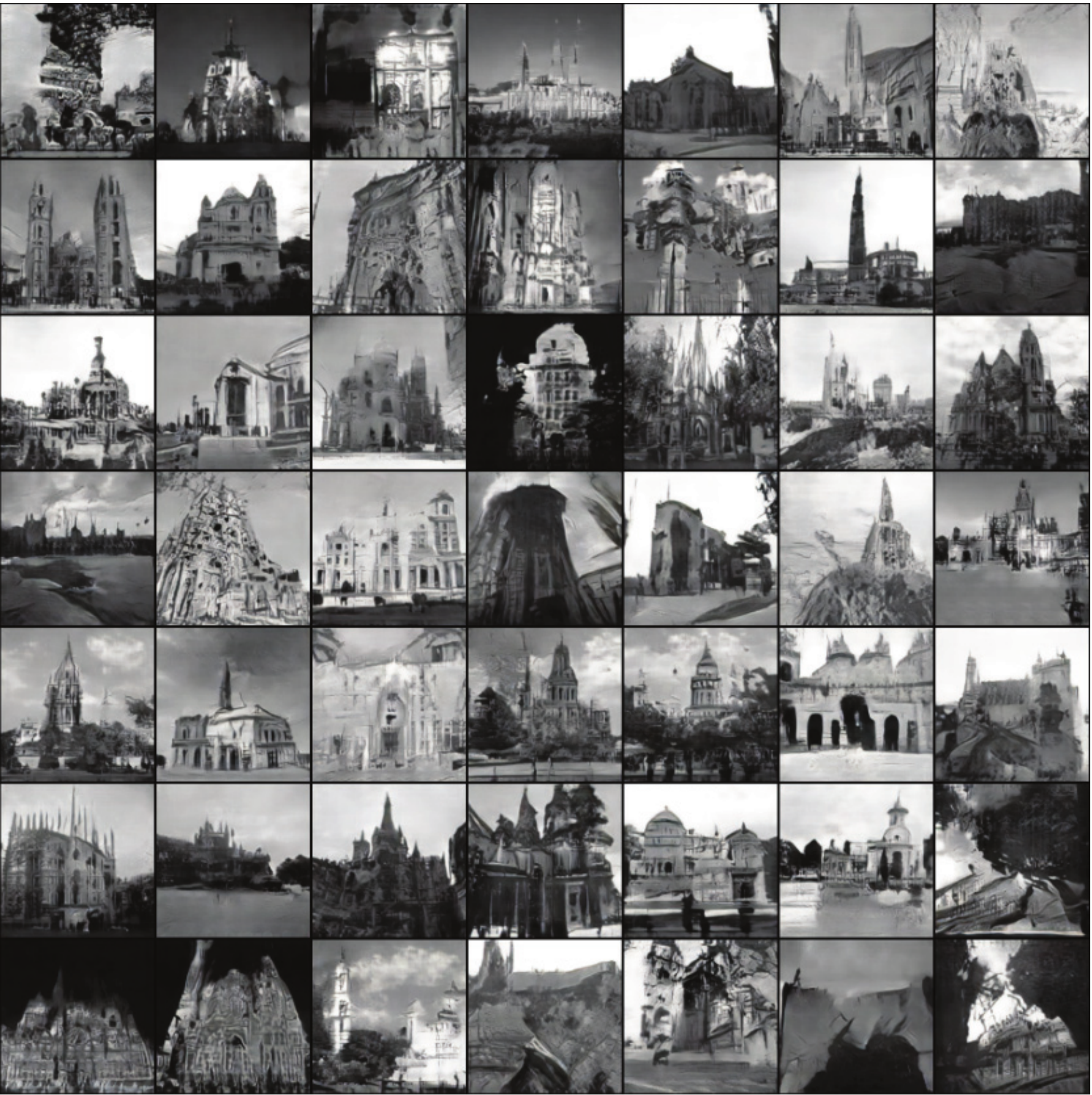}
		}
		\caption{
			Randomly generated samples on gray-Cathedral, supplementing Section \ref{sec:EXP_AdaFM_style} in the main manuscript. Better viewed with zoom in. 
		}
		\label{fig:gray_cathedral}
	\end{center}
\end{figure*}

Concerning medical/biological applications with gray-scale images, we conduct experiments on a gray-scale variant of Cathedral, termed gray-Cathedral. The randomly generated samples are shown in Figure \ref{fig:gray_cathedral}.
Obviously, without AdaFM, worse (blurry and messy) details are observed in the generated images, likely because of the mismatched correlation among channels between source and target domains.

\section{Contributions of the Proposed AdaFM and the Transferred General Part}
\label{secapp:contri_AdaFM}

\begin{figure*}[h]
	\begin{center}
		\includegraphics[width=1.9\columnwidth]{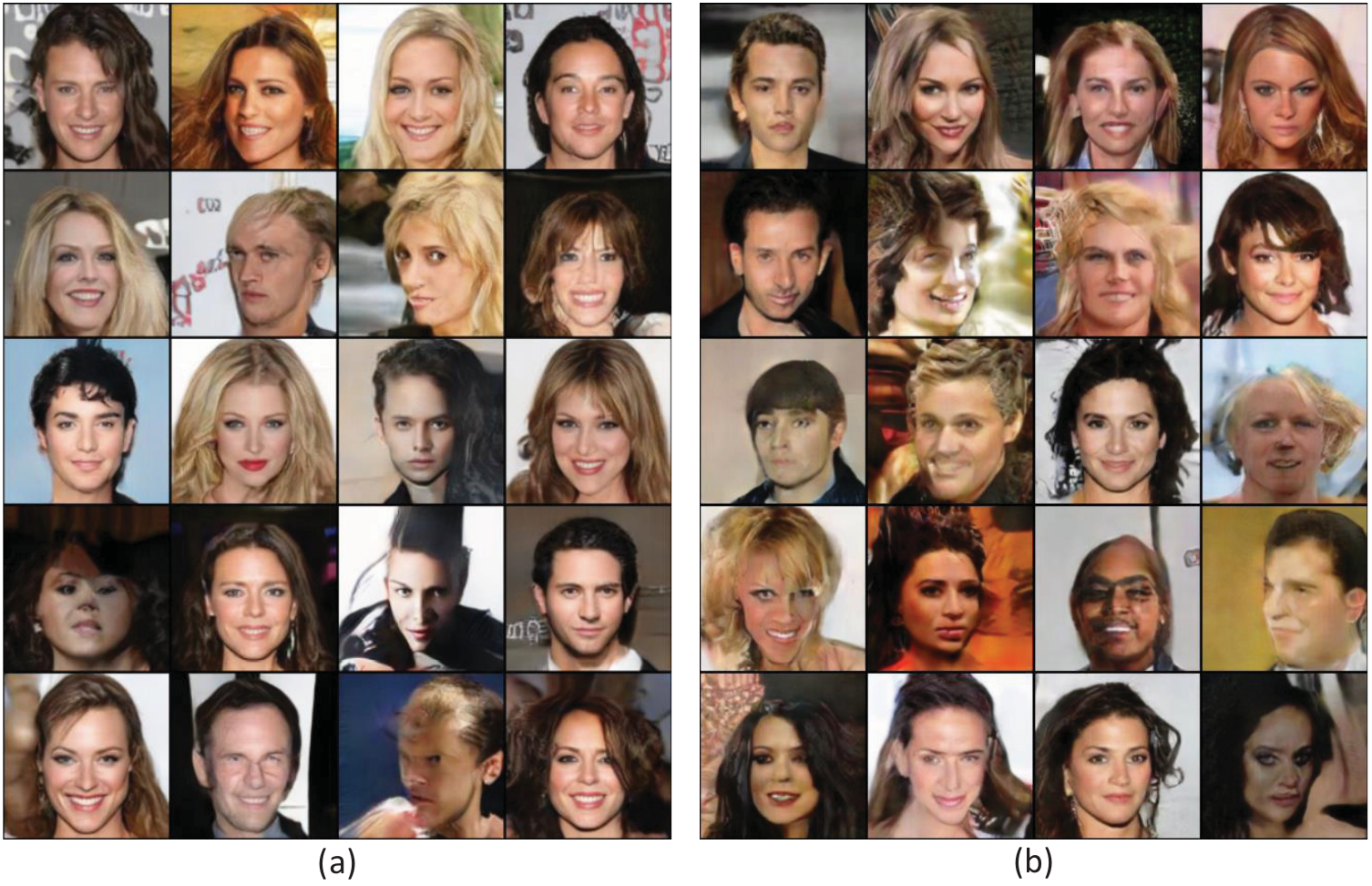}
		\caption{
			Randomly generated samples from our method and SmallHead on CelebA. (a) Our (with AdaFM); (b) SmallHead (without AdaFM). 
		}
		\label{fig:CELEBA_our_adafm}
	\end{center}
\end{figure*}

\begin{figure*}[h]
	\begin{center}
		\includegraphics[width=1.9\columnwidth]{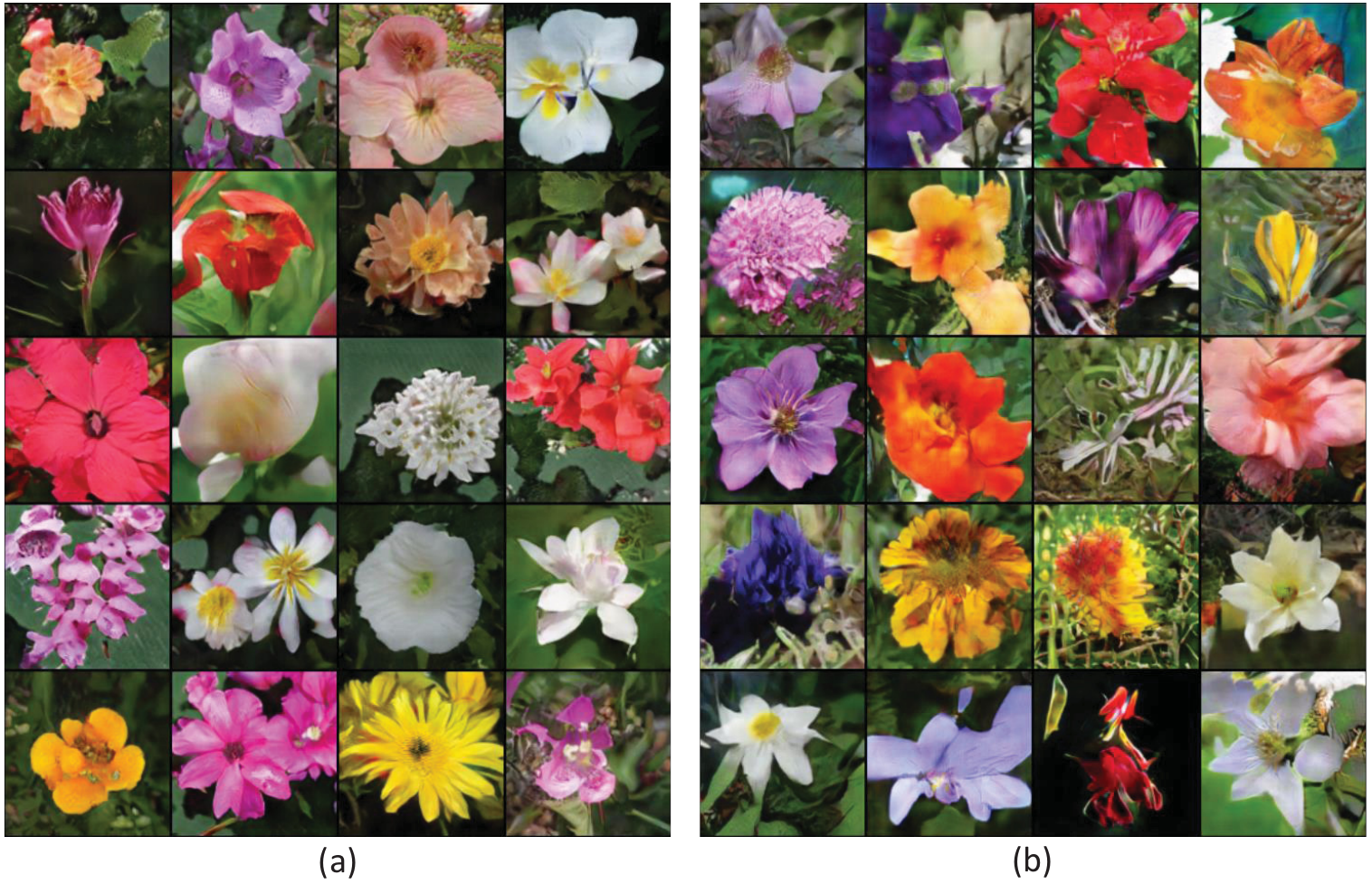}
		\caption{
			Randomly generated samples from our method and SmallHead on Flowers. (a) Our (with AdaFM); (b) SmallHead (without AdaFM).
		}
		\label{fig:Flower_our_adafm}
	\end{center}
\end{figure*}

\begin{figure*}[h]
	\begin{center}
		\includegraphics[width=1.9\columnwidth]{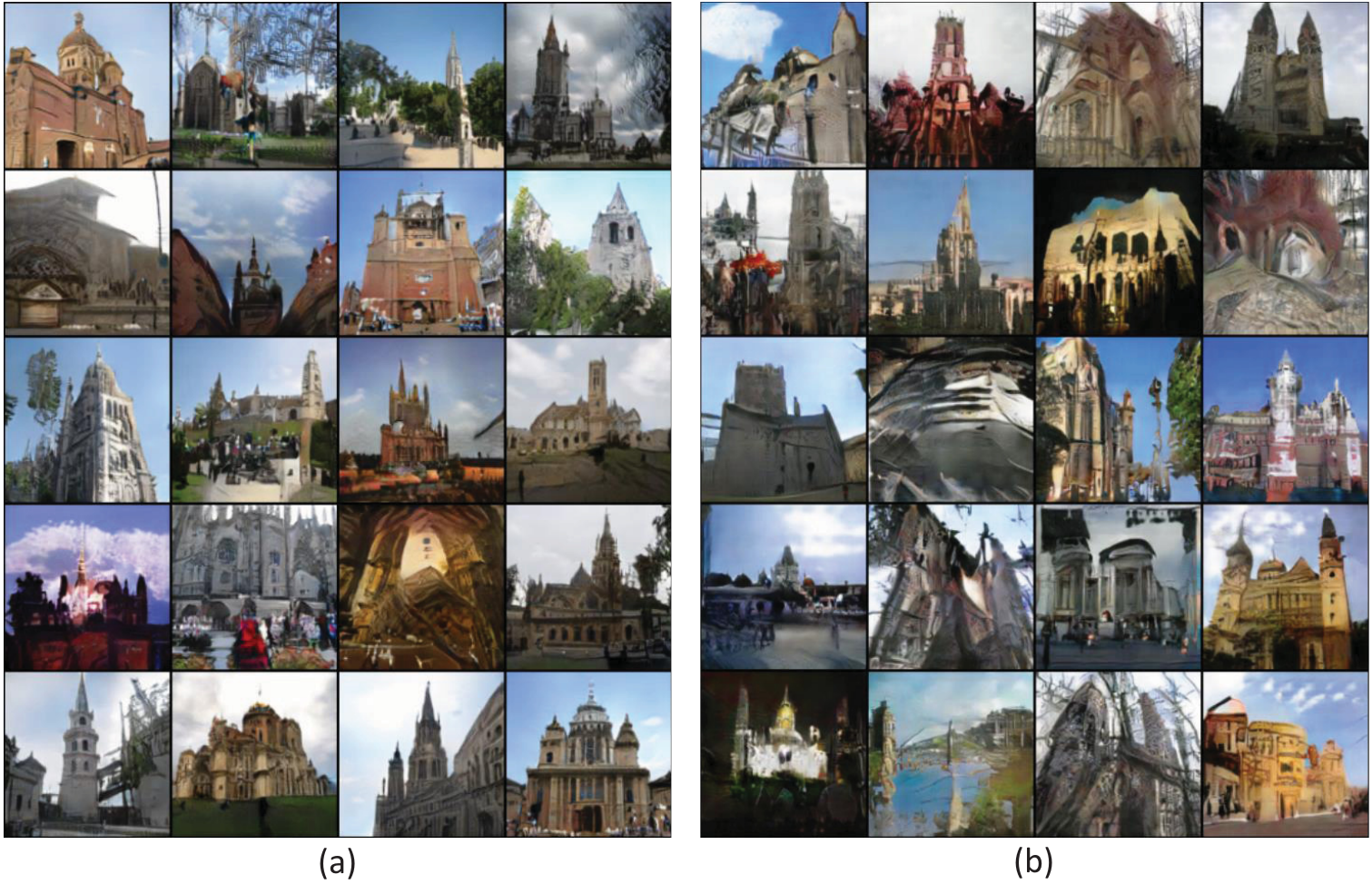}
		\caption{
			Randomly generated samples from our method and SmallHead on Cathedral. (a) Our (with AdaFM); (b) SmallHead (without AdaFM). 
		}
		\label{fig:Cathedral_our_adafm}
	\end{center}
\end{figure*}

To demonstrate the contribution of the proposed AdaFM, the randomly generated samples from our method (with AdaFM) and SmallHead (without AdaFM) on different target domains are shown in Figures \ref{fig:CELEBA_our_adafm}, \ref{fig:Flower_our_adafm}, and \ref{fig:Cathedral_our_adafm}, respectively.
Note that the only difference between our method and SmallHead is the use of AdaFM.
It's clear that, with AdaFM, our method delivers significantly improved generation quality over the Smallhead.

It is worth noting that on all these perceptually-distinct target datasets (\eg CelebA, Flowers, Cathedral), the proposed SmallHead with the transferred general part has proven to train successfully and delivers diverse and relatively-realistic generations, despite without modulations from AdaFM. 
Such phenomena prove that the G4D2 general part discovered in Section \ref{sec:reuse_Big} of the main manuscript generalizes well to various target domains.

\section{Style Mixing on Flowers and CelebA}
\label{sec:style_mixing}

The style-mixing results shown in Figure 12 of the main manuscript are obtained as follows. 

Following \cite{Karras_2019_CVPR}, given the generative process of a ``source'' image, we replace its style input of Group $5$\footnote{We choose Group $5$ for example demonstration; one can of course control the input to other groups, or even hierarchically control all of them.} (the arrow on the left of the specific part of our model; see Figure \ref{fig:Network_all}(h)) with that from a ``Destination'' image, followed by propagating through the rest of the generator to generate a new image with mixed style.

\begin{figure*}[!h]
	\begin{center}
		\includegraphics[width=1.9\columnwidth]{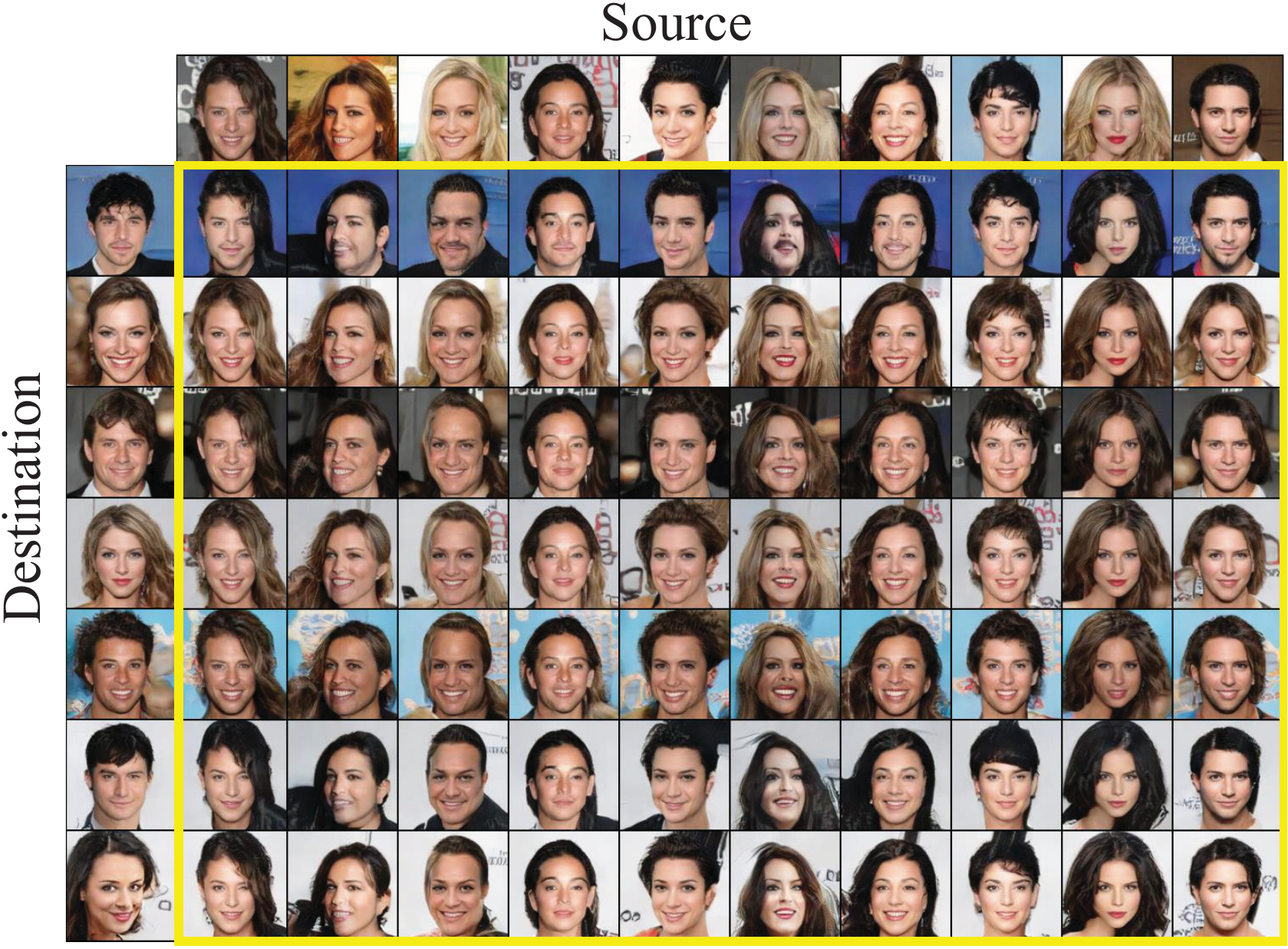}
		\caption{
			Style mixing on CelebA via the tailored specific part of our method.
			The ``Source'' sample controls the identity, posture, and hair type, while the ``Destination'' sample controls the sex, color, and expression.
		}
		\label{fig:mix_style_CELEBA}
	\end{center}
\end{figure*}

A similar style mixing is conducted on CelebA, with the results shown in Figure \ref{fig:mix_style_CELEBA}. We observe that the style inputs from the ``Source'' images control the identity, posture, and hair type, while the style inputs from the ``Destination'' images control the sex, color, and expression.

\end{document}